\documentclass{article} 

\usepackage{amsmath,amsfonts,bm}

\def\eqref#1{equation~\ref{#1}}

\def\1{\bm{1}}

\DeclareMathAlphabet{\mathsfit}{\encodingdefault}{\sfdefault}{m}{sl}
\SetMathAlphabet{\mathsfit}{bold}{\encodingdefault}{\sfdefault}{bx}{n}

\DeclareMathOperator*{\argmax}{arg\,max}

\usepackage{hyperref}

\usepackage[accepted]{icml2020}

\icmltitlerunning{Maximum Likelihood Label Shift with Bias-Corrected Calibration}

\usepackage{graphicx}
\usepackage{subfigure}
\usepackage{url}
\usepackage{microtype}
\usepackage{booktabs}
\usepackage{amsmath}
\usepackage{amssymb} 
\usepackage{amsfonts}

\usepackage{mathtools}
\usepackage{mathrsfs}
\usepackage{multirow}
\usepackage{adjustbox}
\usepackage{placeins}
\usepackage{chngcntr}

\newcommand*{\QED}{\hfill\ensuremath{\blacksquare}}%

\newcommand{\mathbbm}[1]{\text{\usefont{U}{dsrom}{m}{n}#1}}

\allowdisplaybreaks

\begin{document}

\twocolumn[
\icmltitle{Maximum Likelihood with Bias-Corrected Calibration is Hard-To-Beat at Label Shift Adaptation}



\icmlsetsymbol{equal}{*}

\begin{icmlauthorlist}
\icmlauthor{Amr M. Alexandari}{equal,cs}
\icmlauthor{Anshul Kundaje}{cs,gen}
\icmlauthor{Avanti Shrikumar}{equal,cs}
\end{icmlauthorlist}

\icmlaffiliation{cs}{Department of Computer Science, Stanford University}
\icmlaffiliation{gen}{Departments of Genetics, Stanford University}

\icmlcorrespondingauthor{Anshul Kundaje}{anshul@kundaje.net}
\icmlcorrespondingauthor{Avanti Shrikumar}{avanti.shrikumar@gmail.com}

\icmlkeywords{Machine Learning, ICML}

\vskip 0.3in
]



\printAffiliationsAndNotice{\icmlEqualContribution} 

\begin{abstract}
Label shift refers to the phenomenon where the prior class probability $p(y)$ changes between the training and test distributions, while the conditional probability $p(\boldsymbol{x}|y)$ stays fixed. Label shift arises in settings like medical diagnosis, where a classifier trained to predict disease given symptoms must be adapted to scenarios where the baseline prevalence of the disease is different. Given estimates of $p(y|\boldsymbol{x})$ from a predictive model, Saerens et al. proposed an efficient maximum likelihood algorithm to correct for label shift that does not require model retraining, but a limiting assumption of this algorithm is that $p(y|\boldsymbol{x})$ is calibrated, which is not true of modern neural networks. Recently, Black Box Shift Learning (BBSL) and Regularized Learning under Label Shifts (RLLS) have emerged as state-of-the-art techniques to cope with label shift when a classifier does not output calibrated probabilities, but both methods require model retraining with importance weights and neither has been benchmarked against maximum likelihood. Here we (1) show that combining maximum likelihood with a type of calibration we call bias-corrected calibration outperforms both BBSL and RLLS across diverse datasets and distribution shifts, (2) prove that the maximum likelihood objective is concave, and (3) introduce a principled strategy for estimating source-domain priors that improves robustness to poor calibration. This work demonstrates that the maximum likelihood with appropriate calibration is a formidable and efficient baseline for label shift adaptation; notebooks reproducing experiments available \url{https://github.com/kundajelab/labelshiftexperiments}
\end{abstract}

\section{Introduction}

Imagine we train a classifier to predict whether or not a person has a disease based on observed symptoms, and the classifier predicts reliably when deployed in the clinic. Suppose that there is a sudden surge in cases of the disease. During such an outbreak, the probability of persons having the disease given that they show symptoms rises, but the symptoms generated by the disease do not change. How can we adapt the classifier to cope with the difference in the baseline prevalence of the disease?

Formally, let $y$ denote our labels (e.g. whether or not a person is diseased), and let $\boldsymbol{x}$ denote the observed symptoms. Let us denote the joint distribution $(\boldsymbol{x},y)$ before the outbreak (our ``source'' domain) using the letter $\mathbb{P}$, and let us denote the distribution during the outbreak (our ``target'' domain, where we do not have labels) as $\mathbb{Q}$. How can we adapt a classifier trained to estimate $p(y|\boldsymbol{x})$ (the conditional probability in distribution $\mathbb{P}$) so that it can instead estimate  $q(y|\boldsymbol{x})$ (the conditional probability in distribution $\mathbb{Q}$)? Absent assumptions about the nature of the shift between $\mathbb{P}$ and $\mathbb{Q}$, this problem is intractable. However, if the disease generates similar symptoms regardless of spread, we can assume that $p(\boldsymbol{x}|y) = q(\boldsymbol{x}|y)$, and that the shift in the joint distribution $q(\boldsymbol{x},y)$ is due to a shift in the label proportion $q(y)$. Formally, we assume that $q(\boldsymbol{x},y) = p(\boldsymbol{x}|y) q(y)$. This is known as \emph{label shift} or \emph{prior probability shift} \citep{Amos2008-hl}, and it corresponds to anti-causal learning (i.e. predicting the cause $y$ from its effects $\boldsymbol{x}$) \citep{Schoelkopf2012-px}. Anti-causal learning is appropriate for diagnosing diseases given observations of symptoms because diseases cause symptoms.

Given estimates of $p(y)$ and $p(y|\boldsymbol{x})$, \citet{Saerens2002-jh} proposed a simple Expectation Maximization (EM) procedure to estimate $q(y)$ without needing to estimate $p(\boldsymbol{x}|y)$. However, estimates of $p(y|\boldsymbol{x})$ derived from modern neural networks are often poorly calibrated \citep{Guo2017-wk}, and the lack of calibration can decrease the effectiveness of EM. As an alternative, \citet{bbse} developed a technique called Black Box Shift Learning (BBSL) that can work even when the predictions $p(y|\boldsymbol{x})$ are not calibrated. \citet{azizzadenesheli2018regularized} further improved upon BBSL in a technique known as Regularized Learning under Label Shifts (RLLS). Both BBSL and RLLS leverage information in a confusion matrix calculated on a held-out portion of the training set.  To our knowledge, neither BBSL nor RLLS have been benchmarked against EM. Moreover, both BBSL and RLLS require model retraining using importance weighting, which does not work not as well as expected with deep neural networks \citep{importanceweighting}, and RLLS also relies on a regularization hyperparameter. Conversely, EM requires neither retraining nor hyperparameter tuning.

Although the EM approach is limited by the assumption that the predictions $p(y|\boldsymbol{x})$ are calibrated, a number of recent techniques have been proposed to correct for miscalibration of $p(y|\boldsymbol{x})$ using a held-out portion of the training set \citep{Guo2017-wk}. The held-out set can be thought of as analogous to the held-out set used in BBSL and RLLS to calculate a confusion matrix. This suggests a simple yet novel hybrid algorithm for adapting to label shift: first, calibrate predictions using the held-out training set, then perform domain adaptation on the calibrated predictions using EM. In this work, we studied the effectiveness of this hybrid algorithm. More generally, we studied the impact of calibration on domain adaptation to label shift.

\subsection{Our Contributions}

\begin{enumerate}
    \item In experiments on MNIST, CIFAR10/CIFAR100, and Diabetic Retinopathy Detection, we found that EM achieves \textbf{state-of-the-art results} when used with an appropriate type of calibration. Although BBSL and RLLS both benefit from calibration, they did not tend to outperform EM when the probabilities were well-calibrated.
    \item We observed that the popular calibration approach of Temperature Scaling (TS) \citep{Guo2017-wk} does not tend to achieve the best results for adaptation to label shift, possibly owing to large systematic biases in the calibrated probabilities (\textbf{Fig.} \ref{fig:intro_plot}). Best results are obtained with variants of TS containing class-specific bias parameters that can correct for systematic bias.
    \item We make two contributions to the algorithm for maximum likelihood label shift: first, we identify a principled strategy for computing the source-domain priors that improves robustness when the calibrated probabilities have systematic bias. Second, we prove that the likelihood function is concave and bounded; thus, the EM algorithm converges to a global maximum, and standard convex optimizers can be used as an alternative to EM.
\end{enumerate}

\section{Background}

\subsection{Temperature Scaling, Vector Scaling and Expected Calibration Error}

Calibration has a long history in the machine learning literature \citep{degroot1983comparison, Platt1999-qp, zadrozny2001obtaining, zadrozny2002transforming, niculescu2005predicting, kuleshov2015calibrated, ece, kuleshov2016reliable}. In the context of modern neural networks, \citet{Guo2017-wk} showed that Temperature Scaling, a single-parameter variant of Platt Scaling \citep{Platt1999-qp}, was effective at reducing miscalibration. Temperature scaling performs calibration by introducing a temperature parameter $T$ to the logit vector of the softmax. Let $z(\boldsymbol{x}^k)$ represent a vector of the original softmax logits computed on input $\boldsymbol{x}^k$, and let $y$ be a random variable representing the label. With temperature scaling, we have $p(y=i|\boldsymbol{x}^k) = \frac{e^{z(\boldsymbol{x}^k)_i/T}}{\sum_j e^{z(\boldsymbol{x}^k)_j/T}}$, where $T$ is optimized with respect to the Negative Log Likelihood (NLL) on a held-out portion of the training set, such as the validation set. \citet{Guo2017-wk} compared TS to an approach defined as Vector Scaling (VS), where a different scaling parameter was used for each class along with class-specfic bias parameters. Formally, in vector scaling, $p(y=i|\boldsymbol{x}^k) = \frac{e^{(z(\boldsymbol{x}^k)_i W_i) + b_i}}{\sum_j e^{(z(\boldsymbol{x}^k)_j W_j) + b_j}}$. \citet{Guo2017-wk} found that vector scaling had a tendency to perform slightly worse than TS as measured by a metric known as the Expected Calibration Error \citep{ece}. To compute the ECE, the predicted probabilities for the output class are partitioned into $M$ equally spaced bins, and the weighted average of the difference between the bin's accuracy and the bin's confidence is computed, where the weights are determined by the proportion of examples falling in the bin. Formally, $\text{ECE} = \sum_{m=1}^M \frac{|B_m|}{n} |\text{acc}(B_m) - \text{conf}(B_m)|$, where $n$ is the number of samples.



\begin{figure}[!h]
\centering
\includegraphics[width=0.45\textwidth]{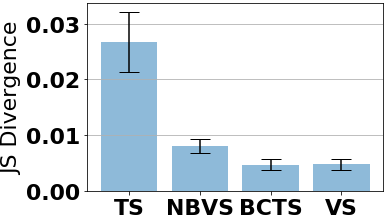}
\caption{\label{fig:intro_plot} \textbf{Temperature Scaling exhibits systematic bias}. On CIFAR10 data, systematic bias was quantified by the JS divergence between the true class label proportions and the average class predictions on a held-out test set drawn from the same distribution as the dataset used for calibration. TS: Temperature Scaling, NBVS: No-Bias Vector Scaling, BCTS: Bias-Corrected Temperature Scaling, VS: Vector Scaling. BCTS and VS had significantly lower systematic bias compared to TS and NBVS. Results are averaged over multiple models and dataset samples (\textbf{Sec. ~\ref{sec:experimentalsetup}}).}
\end{figure}

\subsection{Label Shift Adaptation via Maximum Likelihood}

In a seminal paper on label shift adaptation, \citet{Saerens2002-jh} proposed an Expectation Maximization algorithm for estimating the shift in the class priors between the training and test distributions. Let $\hat{q}^{(s)}(y=i)$ denote the estimate (from EM iteration s) of the prior probability $q(y=i)$ of observing class $i$ in the test set. The algorithm proceeds as follows: first, $\hat{q}^{(0)}(y=i)$ is initialized to be equal to the class priors $\hat{p}(y=i)$ estimated from the training set. Then, the conditional probabilities in the E-step are computed as $\hat{q}^{(s)}(y=i|\boldsymbol{x}_k) =  \frac{\frac{\hat{q}^{(s)}(y=i)}{\hat{p}(y=i)}\hat{p}(y=i|\boldsymbol{x}_k)}{\sum_{j=1}^{m} \frac{\hat{q}^{(s)}(y=j)}{\hat{p}(y=j)}\hat{p}(y=j|\boldsymbol{x}_k)}$, where $m$ is the number of output classes. Finally, the prior estimates in the M-step are updated as $\hat{q}^{(s+1)}(y=i) = \frac{1}{N} \sum_{k=1}^{N} \hat{q}^{(s)}(y=i|\boldsymbol{x}_k)$, where $N$ is the number of examples in the testing set. The E and M steps are iterated until convergence. As there is no need to estimate $p(\boldsymbol{x}|y)$ in any step of the EM procedure, the algorithm can scale to high-dimensional datasets, with a runtime of $O(mN)$ per iteration (independent of the input dimensions). This procedure is motivated by the assumption that $\hat{p}(y=i|\boldsymbol{x}_k)$ is calibrated.

\subsection{Label Shift Adaptation via BBSL \& RLLS}

Following the EM approach of \citet{Saerens2002-jh}, several additional approaches for labels shift adaptation have emerged \citep{chan2005word, storkey2009training, Schoelkopf2012-px, pmlr-v28-zhang13d, bbse, azizzadenesheli2018regularized}. Many of these approaches build estimates $p(x|y)$, which can scale poorly with dataset sizes and underperform on high-dimensional data \citep{bbse}. \citet{bbse} proposed Black-Box Shift Learning (BBSL), which strives to efficiently estimate the weights $[\boldsymbol{w}]_i = \frac{q(y=i)}{p(y=i)}$ even in cases where the prediction model $\hat{p}(y=i|\boldsymbol{x}_k)$ is poorly calibrated or biased. BBSL proceeds as follows: let $f$ be a function that accepts an input and returns the model's predicted class, let $\boldsymbol{x}_k$ denote an example from a held-out portion of the training set, and let $\boldsymbol{x}'_k$ denote an example from the testing set. The empirical estimate of $\boldsymbol{w}$, denoted as $\boldsymbol{\hat{w}}$, is computed as $\boldsymbol{\hat{w}} = \boldsymbol{\hat{C}}_{\hat{y},y}^{-1} \boldsymbol{\hat{u}}_{\hat{y}}$, where $[\boldsymbol{\hat{u}}_{\hat{y}}]_i = \frac{\sum_k \mathbbm{1}\{ f(\boldsymbol{x}'_k) = i\}}{m}$ and $[\boldsymbol{\hat{C}}_{\hat{y},y}]_{ij} = \frac{1}{n} \sum_k \mathbbm{1}\{ f(\boldsymbol{x}_k) = i \text{ and } y_k = j\}$. Because the approach above is not guaranteed to produce positive values for all elements of $\boldsymbol{\hat{w}}$, any negative elements of $\boldsymbol{\hat{w}}$ are set to $0$ after they are estimated. Domain adaptation is then performed by retraining the model on the entire training set distribution with examples upweighted in accordance with $\boldsymbol{\hat{w}}$. \citet{bbse} denote the version of BBSL described above as \textbf{BBSL-hard}. They also compare to a variant that they call \textbf{BBSL-soft}, which they describe as the case where where $f$ outputs probabilities rather than hard classes. We interpreted this to mean $[\boldsymbol{\hat{u}}_{\hat{y}}]_i = \frac{\sum_k f(\boldsymbol{x}'_k)_i}{m}$ and $[\boldsymbol{\hat{C}}_{\hat{y},y}]_{ij} = \frac{1}{n} \sum_k  f(\boldsymbol{x}_k)_i\mathbbm{1}\{y_k = j\}$. \citet{azizzadenesheli2018regularized} further improved upon BBSL by including regularization terms in a technique known as Regularized Learning under Label Shift (RLLS). In our experiments, we compare to BBSL-hard, BBSL-soft, RLLS-hard and RLLS-soft. Regularization hyperparameters for RLLS were set in accordance with the hard-coded values given in the publicly available code provided by the authors at \href{https://github.com/Angela0428/labelshift/blob/5bbe517938f4e3f5bd14c2c105de973dcc2e0917/label_shift.py#L453-L456}{this url}.
Note that BBSL and RLLS both require a portion of the training set to be held out during the initial training phase in order to accurately estimate the confusion matrix $\boldsymbol{\hat{C}_{\hat{y},y}}$; in our experiments involving calibration, we use this same heldout set to calibrate the model.

\section{Methods}



\subsection{No-Bias Vector Scaling and Bias-Corrected Temperature Scaling}

As shown in \textbf{Fig. ~\ref{fig:intro_plot}}, we often found that TS alone resulted in systematically biased estimates of $p(y_i |\boldsymbol{x}^{k})$, while VS, a generalization of TS that contains both class-specific bias terms and class-specific scaling terms, did not exhibit as much systematic bias. Intrigued by this observation, we investigated the performance of two intermediaries between Temperature Scaling and Vector Scaling. The first, which we refer to as No Bias Vector Scaling (NBVS), is equivalent to vector scaling but with all the class-specific bias parameters fixed at zero. The second, which we refer to as Bias-Corrected Temperature Scaling, is equivalent TS Scaling but with the addition of the class-specific bias terms from VS.  As with TS and VS, the parameters are optimized to minimize the NLL on the validation set. Note that in the case of binary classification, the parameterization of BCTS reduces to Platt Scaling \citep{Platt1999-qp}. Thus, BCTS can be viewed as a multi-class generalization of Platt scaling. Given fitted calibration parameters, performing calibration on test data takes $O(mN)$ time where $m$ is the number of output classes and $N$ is the number of datapoints.

\subsection{Defining source-domain priors in the EM}
\label{sec:em_sourcedomain_priors_definition}

The EM algorithm of \citep{Saerens2002-jh} requires the user to provide estimates of the source-domain prior class probabilities $\hat{p}(y=i)$. Let us consider two possible approaches to estimating these probabilities. The first approach, considered in the original paper, is to set $\hat{p}(y=i)$ to the expected value of the binary label $y=i$ over the source domain dataset. A second, less obvious, approach is to set it to the expected value of $\hat{p}(y=i|x)$ over the source domain dataset, formally denoted as $\mathbf{E}_{\boldsymbol{x} \sim p(\boldsymbol{x})}[\hat{p}(y=i|\boldsymbol{x})]$. If $\hat{p}(y=i|x)$ were unbiased, we anticipate that the two approaches would agree. However, depending on the calibration of $\hat{p}(y=i|\boldsymbol{x})$, this may not be the case, bringing us to:

\textbf{Lemma A}: In the absence of domain shift and in the limit of sufficient data, the EM algorithm will converge to the original priors $\hat{p}(y=i)$ if and only if $\hat{p}(y=i) \coloneqq \mathbf{E}_{\boldsymbol{x} \sim p(\boldsymbol{x})}[\hat{p}(y=i|\boldsymbol{x})]$.

\textbf{Proof}: Note that the EM algorithm will converge when $\hat{q}^{(s+1)}(y=i) = \hat{q}^{(s)}(y=i)$. From the M-step, we know that $\hat{q}^{(s+1)}(y=i) = \frac{1}{N} \sum_{k=1}^N \hat{q}^{(s)}(y=i|\boldsymbol{x}_k)$, where the examples $\boldsymbol{x}_k$ are drawn from the target distribution. Substituting the formula for $\hat{q}^{(s)}(y=i|\boldsymbol{x}_k)$ from the E-step, we have $\hat{q}^{(s+1)}(y=i) = \frac{1}{N} \sum_{k=1}^N \frac{ \frac{\hat{q}^{(s)}(y=i)}{\hat{p}(y=i)} \hat{p}(y=i|\boldsymbol{x}_k) }{ \sum_{j=1}^m \frac{\hat{q}^{(s)}(y=j)}{\hat{p}(y=j)} \hat{p}(y=j|\boldsymbol{x}_k) }$. To prove \textbf{sufficiency}, we consider the scenario where $\hat{q}^{(s)}(y=i) = \hat{p}(y=i)$ and check whether convergence is attained. If the samples in the target distribution are drawn from the same distribution as the source, then in the limit of sufficient $N$, the value of $\hat{q}^{(s+1)}(y=1)$ will approach $\mathbf{E}_{x \sim p(x)} \frac{ \frac{1}{1} \hat{p}(y=i|\boldsymbol{x}_k) }{ \sum_{j=1}^m \frac{1}{1} \hat{p}(y=j|\boldsymbol{x}_k)} = \mathbf{E}_{x \sim p(x)} \hat{p}(y=i|\boldsymbol{x}_k)$. As we defined $\hat{p}(y=i)$ to equal $\mathbf{E}_{x \sim p(x)} \hat{p}(y=i|\boldsymbol{x}_k)$, we get $\hat{q}^{(s+1)}(y=1) = \hat{q}^{(s)}(y=1)$, and convergence is attained. To prove \textbf{necessity}, we simply observe that if we had \emph{not} defined $\hat{p}(y=i)$ to equal $\mathbf{E}_{x \sim p(x)} \hat{p}(y=i|\boldsymbol{x}_k)$, then we would not have $\hat{q}^{(s+1)}(y=1) = \hat{q}^{(s)}(y=1)$, and convergence would not be attained. $\QED$

In the absence of domain shift, it is desirable that EM converge to the original priors $\hat{p}(y=i)$. In light of \textbf{Lemma A}, we set $\hat{p}(y=i)$ to the average value of $\hat{p}(y=i|\boldsymbol{x})$ over the source-domain validation set (we use the validation set to avoid the effects of overfitting on the training set; this is the same validation set used for calibration). If we instead compute $\hat{p}(y=i)$ by averaging binary labels in the validation set, we observe poor (even detrimental) performance with EM when the calibration lacks bias correction (\textbf{Tab.} ~\ref{tab:kaggledr_diffemsourcepriors_deltaperf}).

\vspace{-5pt}

\subsection{Convergence of Maximum Likelihood to the Global Optimum}
\label{sec:emunimodal}

We prove that the maximum likelihood function for label shift is concave. As a result, the EM algorithm of \citet{Saerens2002-jh} converges to the maximum likelihood estimate.

\textbf{Lemma B}: the maximum likelihood objective is concave.

\textbf{Proof}: Let $\omega_i$ denote membership in class $i$, and let $q(\omega_i)$ \& $p(\omega_i)$ denote target \& source domain priors.  We seek target-domain priors $q(\boldsymbol{\omega})$ that maximize the log-likelihood $l(\boldsymbol{X}; q(\boldsymbol{\omega}))
    = \sum_k\log\sum_i q(\boldsymbol{x}_k|\omega_i)$. Because $q(\boldsymbol{x}_k|\omega_i)$ is not known explicitly, we rewrite the log likelihood in terms of $p(\omega_i|\boldsymbol{x}_k)$ and $p(\omega_i)$ as follows:
\begin{align}
    l(\boldsymbol{X}; q(\boldsymbol{\omega}))
    &= \sum_k\log\sum_i q(\boldsymbol{x}_k|\omega_i) q(\omega_i) \nonumber\\
    &= \sum_k\log\sum_i p(\boldsymbol{x}_k|\omega_i) q(\omega_i) \label{labelshiftassume}\\
    &= \sum_k\log\sum_i \frac{p(\omega_i|\boldsymbol{x}_k)p(\boldsymbol{x}_k)}{p(\omega_i)} q(\omega_i)\label{bayesrule}\\
    &= \sum_k\log \left(p(\boldsymbol{x}_k) \sum_i \frac{p(\omega_i|\boldsymbol{x}_k)}{p(\omega_i)} q(\omega_i)\right) \nonumber\\
    &= \sum_k\log(p(\boldsymbol{x}_k)) + \log \sum_i \frac{p(\omega_i|\boldsymbol{x}_k)}{p(\omega_i)} q(\omega_i)  \nonumber
\end{align}
where (\ref{labelshiftassume}) follows from the label shift assumptions and (\ref{bayesrule}) follows from Bayes' rule. Now note that the maximization is independent of $p(\boldsymbol{x}_k)$. Using the constant $C$ to denote $\sum_k \log(p(\boldsymbol{x}_k))$, the objective can be written as:
\begin{equation}
\begin{aligned}
\max_{q(\boldsymbol{\omega})} \quad & C + \sum_k\log \sum_i \frac{p(\omega_i|\boldsymbol{x}_k)}{p(\omega_i)} q(\omega_i) \label{eqn:rewrittenemobjective}\\
\textrm{s.t.} \quad & \boldsymbol{1}^T \cdot q(\boldsymbol{\omega}) = 1 \\
  &q(\omega_i)\geq0  \quad \forall i   \\
\end{aligned}
\end{equation}
Both $p(\omega_i|\boldsymbol{x}_k)$ and $p(\omega_i)$ are constants with respect to $q(\boldsymbol{\omega})$. Hence, the objective is a constant plus the sum of logs of linear functions in our decision variable, and the constraints are affine. Therefore, the maximization problem is concave.$\QED$

\textbf{Lemma C}: given that $p(\omega_i) \geq \epsilon$ $\forall i$ (i.e. every class considered has a non-zero probability of occurrence in the source domain), the log likelihood is bounded above by $C +N \log (1/\epsilon)$.

\textbf{Proof}: We give a loose bound. Using the form of the likelihood derived in \textbf{Eqn. \ref{eqn:rewrittenemobjective}}, we have:
\begin{align}
    l(\boldsymbol{X}; q(\boldsymbol{\omega}))
    &= C + \sum_k\log \left( \sum_i \frac{p(\omega_i|\boldsymbol{x}_k)}{p(\omega_i)} q(\omega_i) \right)\\
    &\leq C + \sum_k\log \left( \sum_i p(\omega_i|\boldsymbol{x}_k) \cdot \frac{q(\omega_i)}{\epsilon} \right) \label{assume}\\
    &\leq C + \sum_k\log (1/\epsilon \cdot\sum_i p(\omega_i|\boldsymbol{x}_k))\label{probltone}\\
    &\leq C + N \cdot \log(1/\epsilon) \label{prob}
\end{align}
where (\ref{assume}) is using the given assumption on $p(\omega_i)$, (\ref{probltone}) is using the fact that probabilities are $\le 1$, and (\ref{prob}) is due to the fact that probabilities sum up to one.$\QED$

Note that a similar assumption to $p(\omega_i) \geq \epsilon$ $\forall i$ is adopted by BBSL; if the source-domain probability of a class were $0$, we would simply exclude that class from the optimization.

As a consequence of the likelihood being concave and bounded, standard convex optimization algorithms would converge to the global maximum likelihood estimate. We used the Expectation Maximization (EM) algorithm in this work for ease of implementation. For completeness, we include the convergence proof for EM in Appendix \textbf{~\ref{sec:proof_convergence_stationary}}.

\subsection{Summary of Proposed Algorithm}

For convenience of the reader, we summarize our proposed algorithm for label shift adaptation as follows:

\begin{enumerate}
    \item Given a model $f$ that outputs predicted probabilities, calibrated the predictions of $f$ on a held-out validation using an appropriately strong calibration algorithm. In this work, we observed that Bias-Corrected Temperature Scaling and Vector Scaling work well.
    \item Average the calibrated predictions $\hat{p}(y=i|\boldsymbol{x})$ over this held-out validation set to obtain the estimated source-domain class priors $\hat{p}(y=i)$.
    \item Given samples from the target domain, use the estimated $\hat{p}(y=i)$ and the calibrated predictor $\hat{p}(y=i|\boldsymbol{x})$ to optimize the concave maximum likelihood objective in \textbf{Eqn. \ref{eqn:rewrittenemobjective}} w.r.t. the estimated target-domain class proportions $\hat{q}(y=i)$.
    \item After finding $\hat{q}(y=i)$, compute the adapted predictions for the target domain as follows (this is similar to the E-step of EM, and follows from Bayes' rule; a derivation can be found in \citet{Saerens2002-jh}): \begin{equation}
        \hat{q}(y=i|\boldsymbol{x}) = \frac{\frac{\hat{q}(y=i)}{\hat{p}(y=i)} \hat{p}(y=i|\boldsymbol{x})}{\sum_{j=1}^m \frac{\hat{q}(y=j)}{\hat{p}(y=j)} \hat{p}(y=j|\boldsymbol{x})  }
    \end{equation}
    \label{eqn:adaptedpreds}
\end{enumerate}

\subsection{Metrics for evaluating adaptation to label shift}
\label{sec:metrics_for_labelshift}

The first metric we consider is the mean squared error in the true weights compared to the estimated weights \citep{azizzadenesheli2018regularized, bbse}. Let us denote the true target-domain prior as $q(y=i)$ and the true source domain prior as $p(y=i)$. The true class weights are defined as $\boldsymbol{w}_i \coloneqq q(y=i)/p(y=i)$. Both BBSL and RLLS directly output estimated weights $\boldsymbol{\hat{w}}_i$. For Max Likelihood, the weights can be obtained by dividing the estimated target-domain priors $\hat{q}(y=i)$ by the source-domain priors $\hat{p}(y=i)$ (where the source priors are computed as described in Sec. ~\ref{sec:em_sourcedomain_priors_definition}). The mean squared error of the weights is then simply $\frac{1}{N} \sum_i (\boldsymbol{\hat{w}}_i - \boldsymbol{w}_i)^2$, where $N$ is the number of classes. 

The second metric we consider is the improvement in accuracy of the domain-adapted model predictions relative to using the original model predictions. Given the ratio $\hat{q}(y=i)/\hat{p}(y=i)$, the adapted model predictions can be computed as in \textbf{Eqn. \ref{eqn:adaptedpreds}}, similar to the E-step of EM. For Max Likelihood, we use these adapted predictions to assess accuracy. By contrast, in both the BBSL and RLLS papers, model retraining was performed to obtain adapted predictions. Due to computational constraints, as well as recent observations that retraining deep neural networks using importance weights does not work as well as expected \citep{importanceweighting}, we did not perform model retraining. Thus, we use the MSE of importance weights to compare Max Likelihood to BBSL and RLLS, and use accuracy only for comparing the impact of different calibration algorithms on the Maximum Likelihood estimate\footnote{Technically speaking, we could use the class ratios produced by BBSL and RLLS in conjunction with the E-step update to obtain adapted predictions without model retraining; however, when the predictions are not calibrated, this adaptation can sometimes \emph{decrease} test-set accuracy. To avoid potentially misleading the reader into thinking BBSL and RLLS can harm test-set accuracy, we did not report the results of such adaptation. However, the overall performance trends were similar, in that EM with bias-corrected calibration performed the best.}.

\section{Results}
\label{sec:Experiments}
\FloatBarrier

\begin{table*}[!h]
\adjustbox{max width=\textwidth}{
  \centering
  \begin{tabular}{ c | c | c c c | c c c | c c c}
    \multirow{2}{*}{\begin{tabular}{c}\textbf{Shift} \\ \textbf{Estimator} \end{tabular}} & \multirow{2}{*}{\begin{tabular}{c}\textbf{Calibration} \\ \textbf{Method} \end{tabular}} & \multicolumn{3}{| c}{$\alpha=0.1$} & \multicolumn{3}{| c}{$\alpha=1.0$} & \multicolumn{3}{| c}{$\alpha=10$}\\ 
    \cline{3-11}
    & & $n$=2000 & $n$=4000 & $n$=8000 & $n$=2000 & $n$=4000 & $n$=8000 & $n$=2000 & $n$=4000 & $n$=8000\\
    \hline
    \hline
    EM & None & 0.01049; 4.0 & 0.00763; 4.0 & 0.00795; 4.0 & 0.00283; 4.0 & 0.00202; 4.0 & 0.00147; 4.0 & 0.00191; 2.0 & 0.001; 2.0 & 0.0006; 2.0\\
    BBSL-hard & None & 0.00443; 3.0 & 0.00214; 2.0 & 0.00125; 2.0 & 0.00266; 2.0 & 0.00135; 3.0 & 0.00074; 2.0 & 0.00244; 3.0 & 0.00123; 3.0 & 0.00055; 2.5\\
    BBSL-soft & None & \textbf{0.00309; 1.0} & \textbf{0.00133; 1.0} & \textbf{0.00092; 1.0} & \textbf{0.00188; 1.0} & \textbf{0.00108; 1.0} & \textbf{0.00057; 1.0} & 0.00187; 1.0 & \textbf{0.0009; 1.0} & \textbf{0.00047; 1.0}\\
    RLLS-hard & None & 0.00405; 2.0 & 0.00209; 2.0 & 0.00124; 2.0 & 0.00265; 2.5 & 0.00135; 2.0 & 0.00074; 3.0 & 0.00244; 3.0 & 0.00123; 3.0 & 0.00055; 3.0\\
    RLLS-soft & None & \textbf{0.00316; 1.0} & \textbf{0.00127; 1.0} & \textbf{0.0009; 1.0} & \textbf{0.00183; 1.0} & 0.00108; 1.0 & 0.00057; 1.0 & \textbf{0.00187; 1.0} & \textbf{0.0009; 1.0} & \textbf{0.00047; 1.0}\\
    \hline
    \hline
    EM & TS & 0.00944; 2.0 & 0.00705; 2.0 & 0.00693; 2.0 & 0.00262; 2.0 & 0.00192; 2.0 & 0.00148; 2.0 & 0.00195; 2.0 & 0.00101; 2.0 & 0.00058; 2.0\\
    BBSL-soft & TS & \textbf{0.0031; 1.0} & \textbf{0.00132; 1.0} & \textbf{0.00091; 1.0} & \textbf{0.00191; 1.0} & 0.00102; 1.0 & \textbf{0.0006; 0.0} & 0.00179; 1.0 & \textbf{0.00095; 1.0} & \textbf{0.00048; 1.0}\\
    RLLS-soft & TS & \textbf{0.00312; 1.0} & \textbf{0.00125; 1.0} & \textbf{0.0009; 1.0} & \textbf{0.00186; 1.0} & 0.00102; 1.0 & 0.0006; 1.0 & \textbf{0.00179; 1.0} & \textbf{0.00095; 1.0} & \textbf{0.00048; 1.0}\\
    \hline
    \hline
    EM & NBVS & \textbf{0.0014; 0.0} & \textbf{0.00106; 0.0} & \textbf{0.00075; 0.0} & \textbf{0.00133; 0.0} & \textbf{0.00071; 0.0} & \textbf{0.00043; 0.0} & 0.00161; 0.0 & \textbf{0.00079; 0.0} & 0.00047; 0.0\\
    BBSL-soft & NBVS & 0.0028; 1.0 & 0.00137; 1.0 & 0.00098; 1.0 & 0.00175; 1.0 & 0.00096; 1.0 & 0.00061; 1.0 & 0.00179; 1.0 & 0.00081; 1.0 & 0.00048; 1.0\\
    RLLS-soft & NBVS & 0.00275; 1.0 & 0.00131; 1.0 & \textbf{0.00094; 1.0} & 0.00168; 1.0 & 0.00096; 1.0 & 0.00061; 1.0 & 0.00179; 1.0 & 0.00081; 1.0 & 0.00048; 1.0\\
    \hline
    \hline
    EM & BCTS & \textbf{0.00037; 0.0} & \textbf{0.00034; 0.0} & \textbf{0.00022; 0.0} & \textbf{0.00126; 0.0} & \textbf{0.00071; 0.0} & \textbf{0.00043; 0.0} & \textbf{0.00161; 0.0} & \textbf{0.00077; 0.0} & \textbf{0.00047; 0.0}\\
    BBSL-soft & BCTS & 0.00288; 2.0 & 0.00126; 1.0 & 0.00104; 1.0 & 0.00171; 1.0 & 0.00102; 1.0 & 0.0006; 1.0 & 0.00176; 1.0 & 0.00083; 1.0 & 0.00048; 1.0\\
    RLLS-soft & BCTS & 0.00284; 1.0 & 0.00124; 1.0 & 0.00093; 1.0 & 0.00169; 1.0 & 0.00102; 1.0 & 0.0006; 1.0 & 0.00176; 1.0 & 0.00083; 1.0 & 0.00048; 1.0\\
    \hline
    \hline
    EM & VS & \textbf{0.00061; 0.0} & \textbf{0.00031; 0.0} & \textbf{0.0002; 0.0} & \textbf{0.00118; 0.0} & \textbf{0.00067; 0.0} & \textbf{0.0004; 0.0} & \textbf{0.00167; 0.0} & \textbf{0.00076; 0.0} & 0.00045; 0.0\\
    BBSL-soft & VS & 0.00306; 2.0 & 0.00134; 1.0 & 0.00103; 1.0 & 0.00173; 1.0 & 0.00102; 1.0 & 0.00061; 1.0 & 0.00172; 1.0 & 0.00082; 1.0 & 0.00047; 1.0\\
    RLLS-soft & VS & 0.00298; 1.0 & 0.00133; 1.0 & 0.00091; 1.0 & 0.00171; 1.0 & 0.00102; 1.5 & 0.00061; 1.0 & \textbf{0.00172; 1.0} & 0.00082; 1.0 & 0.00047; 1.0\\
  \end{tabular}}
  \caption{\textbf{CIFAR10: Comparison of EM, BBSL and RLLS (dirichlet shift).} Value before the semicolon is the median MSE in the estimated shift weights (as defined in \textbf{Sec. ~\ref{sec:metrics_for_labelshift}}). Value after the semicolon is the median rank of a method relative to the others in the group that use the same calibration. $\alpha$ represents the dirichlet shift parameter (larger $\alpha$ corresponds to less extreme shift), $n$ represents the sample size for both the validation set and the label-shifted test set. A bold value in a group is not significantly different from the best-performing method in the group, as measured by a paired Wilcoxon test at $p < 0.01$. See \textbf{Table ~\ref{tab:cifar10_all_mseweights_even_dirichletshift}} for an equivalent table but with statistical comparisons done across all calibration methods. EM tends to outperform BBSL and RLLS when calibration techniques involving class-specific bias parameters are used.}
  \label{tab:cifar10_emvsbbsevsrlls_mseweights_even_dirichlet}
\end{table*}

\begin{table*}[!h]
\adjustbox{max width=\textwidth}{
  \centering
  \begin{tabular}{ c | c | c c c | c c c}
    \multirow{2}{*}{\begin{tabular}{c}\textbf{Shift} \\ \textbf{Estimator} \end{tabular}} & \multirow{2}{*}{\begin{tabular}{c}\textbf{Calibration} \\ \textbf{Method} \end{tabular}} & \multicolumn{3}{| c}{$\rho=0.01$} & \multicolumn{3}{| c}{$\rho=0.9$}\\ 
    \cline{3-8}
    & & $n$=2000 & $n$=4000 & $n$=8000 & $n$=2000 & $n$=4000 & $n$=8000\\
    \hline
    \hline
    EM & None & 0.00197; 2.0 & 0.00107; 2.0 & 0.00067; 2.0 & \textbf{0.00584; 1.0} & \textbf{0.00305; 0.0} & 0.00316; 2.0\\
    BBSL-hard & None & 0.00197; 3.0 & 0.00122; 4.0 & 0.00079; 4.0 & 0.00897; 3.0 & 0.00689; 4.0 & 0.00608; 4.0\\
    BBSL-soft & None & 0.00161; 1.0 & 0.00099; 1.0 & 0.00057; 1.0 & 0.00662; 2.0 & 0.00379; 1.0 & \textbf{0.00298; 1.0}\\
    RLLS-hard & None & 0.00197; 3.0 & 0.00122; 3.0 & 0.00079; 3.0 & 0.00868; 2.5 & 0.00676; 3.0 & 0.00593; 3.0\\
    RLLS-soft & None & 0.00161; 0.5 & \textbf{0.00099; 1.0} & \textbf{0.00057; 0.0} & \textbf{0.00622; 1.0} & 0.00379; 1.0 & \textbf{0.00298; 1.0}\\
    \hline
    \hline
    EM & TS & 0.00162; 2.0 & \textbf{0.00084; 0.0} & 0.00054; 0.0 & \textbf{0.00318; 0.0} & \textbf{0.00155; 0.0} & \textbf{0.00085; 0.0}\\
    BBSL-soft & TS & 0.00161; 1.0 & 0.00091; 1.0 & 0.00054; 2.0 & 0.00659; 2.0 & 0.00395; 1.0 & 0.00339; 1.0\\
    RLLS-soft & TS & 0.00161; 1.0 & \textbf{0.00091; 1.0} & 0.00054; 1.0 & 0.00627; 1.0 & 0.00393; 1.0 & 0.00339; 1.0\\
    \hline
    \hline
    EM & NBVS & \textbf{0.00158; 0.0} & \textbf{0.00076; 0.0} & \textbf{0.00054; 0.0} & \textbf{0.00112; 0.0} & \textbf{0.00059; 0.0} & \textbf{0.00038; 0.0}\\
    BBSL-soft & NBVS & 0.00162; 1.0 & 0.00088; 2.0 & 0.00057; 2.0 & 0.00607; 2.0 & 0.00385; 1.0 & 0.00349; 1.0\\
    RLLS-soft & NBVS & 0.00162; 1.0 & 0.00088; 1.0 & 0.00057; 1.0 & 0.006; 1.0 & 0.0038; 1.0 & 0.00349; 2.0\\
    \hline
    \hline
    EM & BCTS & \textbf{0.00153; 0.0} & \textbf{0.00076; 0.0} & \textbf{0.00052; 0.0} & \textbf{0.0006; 0.0} & \textbf{0.00037; 0.0} & \textbf{0.00028; 0.0}\\
    BBSL-soft & BCTS & 0.00158; 1.0 & 0.00087; 2.0 & 0.00055; 2.0 & 0.00623; 2.0 & 0.00383; 1.0 & 0.00338; 1.0\\
    RLLS-soft & BCTS & \textbf{0.00158; 1.0} & 0.00087; 1.0 & 0.00055; 1.0 & 0.00603; 1.0 & 0.00378; 2.0 & 0.00338; 2.0\\
    \hline
    \hline
    EM & VS & \textbf{0.00155; 0.0} & \textbf{0.00076; 0.0} & \textbf{0.00052; 0.0} & \textbf{0.00067; 0.0} & \textbf{0.00042; 0.0} & \textbf{0.00033; 0.0}\\
    BBSL-soft & VS & 0.00171; 1.0 & 0.00086; 2.0 & 0.00056; 2.0 & 0.00641; 2.0 & 0.00407; 2.0 & 0.00393; 1.0\\
    RLLS-soft & VS & 0.00171; 1.0 & 0.00086; 1.0 & 0.00056; 1.0 & 0.00626; 1.0 & 0.00405; 1.0 & 0.00393; 2.0\\
  \end{tabular}}
  \caption{\textbf{MNIST: Comparison of EM, BBSL and RLLS (``tweak-one'' shift)}. Value before the semicolon is the median MSE in the estimated shift weights. Value after semicolon is the median rank of a method relative to others in the group that use the same calibration. A bold value in a group is not significantly different from the best-performing method in the group, as measured by a paired Wilcoxon test at $p < 0.01$. See \textbf{Table ~\ref{tab:MNIST_all_mseweights_even_tweakoneshift}} for an equivalent table but with statistical comparisons done across all calibration methods.  EM tends to outperform BBSL and RLLS when calibration techniques involving class-specific bias parameters are used.}
  \label{tab:MNIST_emvsbbsevsrlls_mseweights_even_tweakone}
\end{table*}

\begin{table*}[!h]
\adjustbox{max width=\textwidth}{
  \centering
  \begin{tabular}{ c | c | c c c | c c c | c c c}
    \multirow{2}{*}{\begin{tabular}{c}\textbf{Shift} \\ \textbf{Estimator} \end{tabular}} & \multirow{2}{*}{\begin{tabular}{c}\textbf{Calibration} \\ \textbf{Method} \end{tabular}} & \multicolumn{3}{| c}{$\alpha=0.1$} & \multicolumn{3}{| c}{$\alpha=1.0$} & \multicolumn{3}{| c}{$\alpha=10.0$}\\ 
    \cline{3-11}
    & & $n$=7000 & $n$=8500 & $n$=10000 & $n$=7000 & $n$=8500 & $n$=10000 & $n$=7000 & $n$=8500 & $n$=10000\\
    \hline
    \hline
    EM & None & 1.80113; 4.0 & 1.67187; 4.0 & 1.7157; 4.0 & 0.55795; 4.0 & 0.54112; 4.0 & 0.55955; 4.0 & 0.3896; 4.0 & 0.37585; 4.0 & 0.36337; 4.0\\
    BBSL-hard & None & 0.83095; 3.0 & 0.68099; 3.0 & 0.58656; 3.0 & 0.33542; 3.0 & 0.27998; 3.0 & 0.24659; 3.0 & 0.25694; 3.0 & 0.21736; 3.0 & 0.20637; 3.0\\
    BBSL-soft & None & 0.5279; 2.0 & 0.47521; 2.0 & 0.44263; 2.0 & 0.23637; 2.0 & 0.21324; 2.0 & 0.18605; 2.0 & 0.18842; 2.0 & 0.16108; 2.0 & 0.14018; 2.0\\
    RLLS-hard & None & 0.4577; 1.0 & 0.38413; 1.0 & 0.37675; 1.0 & 0.20285; 1.0 & 0.16017; 1.0 & 0.14228; 1.0 & 0.16412; 1.0 & 0.13842; 1.0 & 0.12127; 1.0\\
    RLLS-soft & None & \textbf{0.33882; 1.0} & \textbf{0.2728; 0.0} & \textbf{0.2878; 0.0} & \textbf{0.13859; 0.0} & \textbf{0.12697; 0.0} & \textbf{0.10567; 0.0} & \textbf{0.12528; 0.0} & \textbf{0.11011; 0.0} & \textbf{0.10056; 0.0}\\
    \hline
    \hline
    EM & TS & 0.41127; 1.0 & 0.34963; 1.0 & 0.30451; 1.0 & 0.23803; 2.0 & 0.21715; 2.0 & 0.19781; 2.0 & 0.16408; 2.0 & 0.14477; 2.0 & 0.13852; 2.0\\
    BBSL-soft & TS & 0.40279; 1.0 & 0.34713; 1.0 & 0.33012; 1.0 & 0.18774; 1.0 & 0.15665; 1.0 & 0.12639; 1.0 & 0.13409; 1.0 & 0.10951; 1.0 & 0.09703; 1.0\\
    RLLS-soft & TS & 0.27236; 1.0 & 0.22426; 1.0 & 0.22697; 1.0 & \textbf{0.12255; 0.0} & \textbf{0.10639; 0.0} & \textbf{0.08221; 0.0} & \textbf{0.11395; 0.0} & \textbf{0.08803; 0.0} & \textbf{0.07868; 0.0}\\
    \hline
    \hline
    EM & NBVS & \textbf{0.1637; 0.0} & \textbf{0.15904; 0.0} & \textbf{0.14348; 0.0} & \textbf{0.1042; 0.0} & \textbf{0.10523; 0.0} & \textbf{0.10777; 1.0} & \textbf{0.10122; 0.0} & \textbf{0.09874; 1.0} & 0.09729; 2.0\\
    BBSL-soft & NBVS & 0.40856; 2.0 & 0.33122; 2.0 & 0.29594; 2.0 & 0.17545; 2.0 & 0.1409; 2.0 & 0.11336; 2.0 & 0.13601; 2.0 & 0.113; 1.0 & 0.09731; 1.0\\
    RLLS-soft & NBVS & 0.2797; 1.0 & 0.21465; 1.0 & 0.23037; 1.0 & \textbf{0.11469; 1.0} & \textbf{0.09903; 1.0} & \textbf{0.08341; 1.0} & \textbf{0.11303; 1.0} & \textbf{0.08868; 1.0} & \textbf{0.07531; 0.0}\\
    \hline
    \hline
    EM & BCTS & \textbf{0.1101; 0.0} & \textbf{0.10824; 0.0} & \textbf{0.11636; 0.0} & \textbf{0.08838; 0.0} & \textbf{0.09102; 0.0} & \textbf{0.0876; 1.0} & \textbf{0.09207; 0.0} & \textbf{0.08707; 1.0} & 0.08781; 2.0\\
    BBSL-soft & BCTS & 0.40887; 2.0 & 0.32629; 2.0 & 0.30435; 2.0 & 0.17923; 2.0 & 0.14756; 2.0 & 0.11536; 2.0 & 0.13698; 2.0 & 0.11267; 2.0 & 0.09748; 1.0\\
    RLLS-soft & BCTS & 0.26594; 1.0 & 0.2305; 1.0 & 0.23604; 1.0 & 0.11498; 1.0 & \textbf{0.09755; 1.0} & \textbf{0.08309; 1.0} & 0.1085; 1.0 & \textbf{0.08672; 1.0} & \textbf{0.07252; 0.0}\\
    \hline
    \hline
    EM & VS & \textbf{0.09485; 0.0} & \textbf{0.08792; 0.0} & \textbf{0.0862; 0.0} & \textbf{0.07684; 0.0} & \textbf{0.07922; 0.0} & \textbf{0.07773; 0.0} & \textbf{0.09387; 0.0} & \textbf{0.08796; 1.0} & 0.08981; 2.0\\
    BBSL-soft & VS & 0.43085; 2.0 & 0.3216; 2.0 & 0.30221; 2.0 & 0.16715; 2.0 & 0.13847; 2.0 & 0.11169; 2.0 & 0.13075; 2.0 & 0.10971; 1.5 & 0.09433; 1.0\\
    RLLS-soft & VS & 0.28941; 1.0 & 0.21363; 1.0 & 0.22405; 1.0 & 0.11627; 1.0 & 0.09894; 1.0 & \textbf{0.08267; 1.0} & \textbf{0.11426; 1.0} & \textbf{0.08623; 1.0} & \textbf{0.07414; 0.5}\\
  \end{tabular}}
  \caption{\textbf{CIFAR100: Comparison of EM, BBSL and RLLS (dirichlet shift).} Value before the semicolon is the median MSE in the estimated shift weights. Value after the semicolon is the median rank of a method relative to the others in the group that use the same calibration. A bold value in a group is not significantly different from the best-performing method in the group, as measured by a paired Wilcoxon test at $p < 0.01$. See \textbf{Table ~\ref{tab:cifar100_all_mseweights_even_dirichletshift}} for an equivalent table but with statistical comparisons done across all calibration methods. EM tends to outperform BBSL and RLLS when calibration techniques involving class-specific bias parameters are used.}
  \label{tab:cifar100_emvsbbsevsrlls_mseweights_even_dirichlet}
\end{table*}

\begin{table*}[!h]
\adjustbox{max width=\textwidth}{
  \centering
  \begin{tabular}{ c | c | c c c | c c c}
    \multirow{2}{*}{\begin{tabular}{c}\textbf{Shift} \\ \textbf{Estimator} \end{tabular}} & \multirow{2}{*}{\begin{tabular}{c}\textbf{Calibration} \\ \textbf{Method} \end{tabular}} & \multicolumn{3}{| c}{$\rho=0.5$} & \multicolumn{3}{| c}{$\rho=0.9$}\\ 
    \cline{3-8}
    & & $n$=500 & $n$=1000 & $n$=1500 & $n$=500 & $n$=1000 & $n$=1500\\
    \hline
    \hline
    EM & None & \textbf{0.35073; 0.0} & \textbf{0.30498; 0.0} & \textbf{0.19709; 0.0} & 0.0899; 2.0 & 0.0596; 2.0 & 0.05666; 3.0\\
    BBSL-hard & None & 1.83879; 4.0 & 1.4584; 4.0 & 0.81962; 4.0 & 0.31856; 4.0 & 0.10582; 4.0 & 0.0491; 3.0\\
    BBSL-soft & None & 1.4041; 2.0 & 0.71345; 2.0 & 0.43408; 2.0 & 0.10441; 2.5 & 0.04494; 2.0 & 0.02562; 2.0\\
    RLLS-hard & None & 1.07889; 2.0 & 1.08159; 3.0 & 0.74669; 3.0 & 0.05118; 1.0 & \textbf{0.02865; 1.0} & 0.02458; 1.0\\
    RLLS-soft & None & 0.91948; 2.0 & 0.63483; 1.0 & 0.41057; 1.0 & \textbf{0.03602; 1.0} & \textbf{0.0225; 1.0} & \textbf{0.02351; 1.0}\\
    \hline
    \hline
    EM & TS & \textbf{0.47048; 0.0} & \textbf{0.29683; 0.0} & \textbf{0.20944; 0.0} & 0.08077; 1.0 & 0.05537; 2.0 & 0.05268; 2.0\\
    BBSL-soft & TS & 1.26587; 2.0 & 0.57077; 2.0 & 0.38372; 2.0 & 0.09552; 1.5 & 0.03859; 1.0 & 0.02165; 1.0\\
    RLLS-soft & TS & 0.80812; 1.0 & 0.56293; 1.0 & 0.37624; 1.0 & \textbf{0.0352; 0.0} & \textbf{0.02457; 0.0} & \textbf{0.02054; 0.5}\\
    \hline
    \hline
    EM & NBVS & \textbf{0.51515; 0.0} & \textbf{0.33256; 0.0} & \textbf{0.24066; 0.0} & 0.09137; 1.5 & 0.08415; 2.0 & 0.08002; 2.0\\
    BBSL-soft & NBVS & 1.62832; 2.0 & 0.70875; 2.0 & 0.45962; 2.0 & 0.09014; 1.0 & 0.03608; 1.0 & \textbf{0.01769; 1.0}\\
    RLLS-soft & NBVS & 0.77171; 1.0 & 0.56492; 1.0 & 0.40272; 1.0 & \textbf{0.04301; 1.0} & \textbf{0.03258; 1.0} & \textbf{0.02541; 0.5}\\
    \hline
    \hline
    EM & BCTS & \textbf{0.40245; 0.0} & \textbf{0.2475; 0.0} & \textbf{0.19461; 0.0} & \textbf{0.02267; 0.0} & \textbf{0.01354; 0.0} & \textbf{0.01043; 0.0}\\
    BBSL-soft & BCTS & 1.40016; 2.0 & 0.52628; 2.0 & 0.46372; 2.0 & 0.05524; 1.0 & 0.02529; 1.0 & 0.01419; 1.0\\
    RLLS-soft & BCTS & 0.87398; 1.0 & 0.50668; 1.0 & 0.42255; 1.0 & \textbf{0.0384; 1.0} & 0.02826; 1.0 & 0.02245; 1.0\\
    \hline
    \hline
    EM & VS & \textbf{0.53081; 0.0} & \textbf{0.26442; 0.0} & \textbf{0.21641; 0.0} & \textbf{0.02388; 0.0} & \textbf{0.01307; 0.0} & \textbf{0.00953; 0.0}\\
    BBSL-soft & VS & 1.36527; 2.0 & 0.50393; 2.0 & 0.48291; 1.5 & 0.04678; 1.0 & 0.02485; 1.0 & 0.01543; 1.0\\
    RLLS-soft & VS & 0.82992; 1.0 & 0.47802; 1.0 & 0.42028; 1.0 & \textbf{0.04096; 1.0} & 0.02948; 1.0 & 0.022; 1.0\\
  \end{tabular}}
  \caption{\textbf{Kaggle Diabetic Retinopathy: Comparison of EM, BBSL and RLLS.} $\rho$ represents proportion of healthy examples in shifted domain; source domain has $\rho=0.73$. Value before semicolon is the median MSE in the estimated shift weights. Value after the semicolon is the median rank of a method relative to others in the group that use the same calibration. A bold value in a group is not significantly different from the best-performing method in the group (paired Wilcoxon test at $p < 0.01$). See \textbf{Table ~\ref{tab:kaggledr_all_mseweights_even}} for an equivalent table but with statistical comparisons done across all calibration methods. EM tends to outperform BBSL and RLLS when calibration techniques involving class-specific bias parameters are used.}
  \label{tab:kaggledr_emvsbbsevsrlls_mseweights_even}
\end{table*}

\begin{table*}[!h]
\adjustbox{max width=\textwidth}{
  \centering
  \begin{tabular}{ c | c | c c c | c c c | c c c}
    \multirow{2}{*}{\begin{tabular}{c}\textbf{Shift} \\ \textbf{Estimator} \end{tabular}} & \multirow{2}{*}{\begin{tabular}{c}\textbf{Calibration} \\ \textbf{Method} \end{tabular}} & \multicolumn{3}{| c}{$\alpha=0.1$} & \multicolumn{3}{| c}{$\alpha=1.0$} & \multicolumn{3}{| c}{$\alpha=10$}\\ 
    \cline{3-11}
    & & $n$=2000 & $n$=4000 & $n$=8000 & $n$=2000 & $n$=4000 & $n$=8000 & $n$=2000 & $n$=4000 & $n$=8000\\
    \hline
    \hline
    EM & None & 5.275; 4.0 & 5.225; 4.0 & 5.319; 4.0 & 1.75; 4.0 & 1.65; 4.0 & 1.669; 4.0 & 0.275; 3.0 & 0.175; 3.0 & 0.25; 4.0\\
    EM & TS & 5.6; 2.0 & 5.513; 3.0 & 5.488; 3.0 & 1.725; 3.0 & 1.713; 3.0 & 1.775; 3.0 & 0.25; 4.0 & 0.188; 3.0 & 0.244; 3.0\\
    EM & NBVS & \textbf{5.85; 2.0} & \textbf{5.725; 2.0} & 5.806; 2.0 & 2.125; 2.0 & 2.013; 2.0 & 2.15; 2.0 & 0.7; 1.0 & 0.775; 1.0 & 0.788; 2.0\\
    EM & BCTS & \textbf{5.825; 2.0} & \textbf{5.775; 1.0} & 5.75; 1.0 & \textbf{2.15; 1.0} & \textbf{2.075; 1.0} & 2.081; 1.0 & \textbf{0.725; 1.0} & \textbf{0.8; 1.0} & \textbf{0.838; 1.0}\\
    EM & VS & 5.625; 2.0 & \textbf{5.675; 1.0} & 5.775; 1.0 & \textbf{2.15; 1.0} & \textbf{2.1; 1.0} & \textbf{2.187; 1.0} & \textbf{0.75; 1.0} & \textbf{0.812; 1.0} & \textbf{0.862; 1.0}\\
  \end{tabular}}
  \caption{\textbf{CIFAR10: Comparison of calibration methods when using EM adaptation to dirichlet shift, with $\Delta$\%accuracy as the metric}. Unlike BBSL and RLLS, the EM algorithm does not rely on retraining to produce domain adapted probabilities. Value before the semicolon is the median change in \%accuracy relative to a baseline of no adaptation. Value after the semicolon is the median rank compared to other methods in the same column. Bold values in a column are not significantly different from the best performing method in the column, as measured by a paired Wilcoxon test at $p \le 0.01$. Calibration techniques involving class-specific bias parameters (namely BCTS and VS) tend to achieve the best performance.}
  \label{tab:cifar10_em_deltaacc_dirichletshift}
\end{table*}

\begin{table*}[!h]
\adjustbox{max width=\textwidth}{
  \centering
  \begin{tabular}{ c | c | c c c | c c c | c c c}
    \multirow{2}{*}{\begin{tabular}{c}\textbf{Shift} \\ \textbf{Estimator} \end{tabular}} & \multirow{2}{*}{\begin{tabular}{c}\textbf{Calibration} \\ \textbf{Method} \end{tabular}} & \multicolumn{3}{| c}{$\alpha=0.1$} & \multicolumn{3}{| c}{$\alpha=1.0$} & \multicolumn{3}{| c}{$\alpha=10.0$}\\ 
    \cline{3-11}
    & & $n$=7000 & $n$=8500 & $n$=10000 & $n$=7000 & $n$=8500 & $n$=10000 & $n$=7000 & $n$=8500 & $n$=10000\\
    \hline
    \hline
    EM & None & 14.136; 4.0 & 14.153; 4.0 & 14.22; 4.0 & 11.971; 4.0 & 12.112; 4.0 & 12.155; 4.0 & 11.779; 4.0 & 11.682; 4.0 & 11.89; 4.0\\
    EM & TS & 24.479; 2.0 & 24.371; 2.0 & 24.855; 2.0 & 21.157; 2.0 & 21.271; 2.0 & 20.99; 2.0 & 20.7; 2.5 & 20.894; 3.0 & 20.915; 2.5\\
    EM & NBVS & 24.079; 2.0 & 24.053; 2.0 & 24.575; 2.0 & 20.986; 2.0 & 21.382; 2.0 & 21.53; 2.0 & 21.036; 2.0 & 20.947; 2.0 & 21.025; 2.0\\
    EM & BCTS & 24.271; 2.0 & 24.112; 1.0 & 24.3; 1.0 & \textbf{21.321; 1.0} & \textbf{21.506; 1.0} & 21.71; 1.0 & \textbf{21.143; 1.0} & \textbf{21.329; 1.0} & \textbf{21.125; 1.0}\\
    EM & VS & \textbf{24.829; 1.0} & \textbf{24.518; 1.0} & 24.51; 1.0 & \textbf{21.3; 1.0} & \textbf{21.629; 1.0} & \textbf{21.93; 1.0} & \textbf{21.221; 1.0} & \textbf{21.282; 1.0} & \textbf{21.21; 1.0}\\
  \end{tabular}}
  \caption{\textbf{CIFAR100: Comparison of calibration methods when using EM adaptation to dirichlet shift, with $\Delta$\%accuracy as the metric}. Unlike BBSL and RLLS, the EM algorithm does not rely on retraining to produce domain adapted probabilities. Value before the semicolon is the median change in \%accuracy relative to a baseline of no adaptation. Value after the semicolon is the median rank compared to other methods in the same column. Bold values in a column are not significantly different from the best performing method in the column, as measured by a paired Wilcoxon test at $p \le 0.01$. Calibration techniques involving class-specific bias parameters (namely BCTS and VS) tend to achieve the best performance.}
  \label{tab:cifar100_em_deltaacc_dirichletshift}
\end{table*}

\begin{table*}[!h]
\adjustbox{max width=\textwidth}{
  \centering
  \begin{tabular}{ c | c | c c c | c c c}
    \multirow{2}{*}{\begin{tabular}{c}\textbf{Shift} \\ \textbf{Estimator} \end{tabular}} & \multirow{2}{*}{\begin{tabular}{c}\textbf{Calibration} \\ \textbf{Method} \end{tabular}} & \multicolumn{3}{| c}{$\rho=0.5$} & \multicolumn{3}{| c}{$\rho=0.9$}\\ 
    \cline{3-8}
    & & $n$=500 & $n$=1000 & $n$=1500 & $n$=500 & $n$=1000 & $n$=1500\\
    \hline
    \hline
    EM & None & 2.0; 3.0 & 2.0; 4.0 & 2.2; 4.0 & 1.4; 4.0 & 1.5; 4.0 & 1.567; 4.0\\
    EM & TS & 2.2; 3.0 & 2.3; 3.0 & 2.667; 3.0 & 1.6; 3.0 & 2.0; 3.0 & 2.133; 3.0\\
    EM & NBVS & 3.5; 2.0 & 4.1; 2.0 & 4.233; 2.0 & 2.2; 2.0 & 2.4; 2.0 & 2.467; 2.0\\
    EM & BCTS & \textbf{3.8; 1.0} & \textbf{4.65; 1.0} & \textbf{4.633; 1.0} & \textbf{3.6; 0.0} & \textbf{3.6; 0.0} & \textbf{3.733; 0.0}\\
    EM & VS & \textbf{3.8; 1.0} & \textbf{4.4; 1.0} & \textbf{4.633; 1.0} & \textbf{3.5; 1.0} & \textbf{3.6; 1.0} & \textbf{3.733; 1.0}\\
  \end{tabular}}
  \caption{\textbf{Kaggle Diabetic Retinopathy: Comparison of calibration methods when using EM adaptation to domain shift, with $\Delta$\%accuracy as the metric.} $\rho$ represents proportion of healthy examples in shifted domain; source distribution has $\rho=0.73$. Unlike BBSL and RLLS, the EM algorithm does not rely on retraining to produce domain adapted probabilities. Value before the semicolon is the median change in \%accuracy relative to a baseline of no adaptation. Value after the semicolon is the median rank compared to other methods in the same column. Bold values in a column are not significantly different from the best performing method in the column, as measured by a paired Wilcoxon test at $p \le 0.01$. Calibration techniques involving class-specific bias parameters (namely BCTS and VS) tend to achieve the best performance. }
  \label{tab:kaggledr_em_deltaacc}
\end{table*}

\subsection{Experimental Setup}
\label{sec:experimentalsetup}

We evaluated the efficacy of BBSL, RLLS and Max Likelihood coupled to different calibration approaches on MNIST, CIFAR10/CIFAR100, and a diabetic retinopathy detection dataset. For our experiments on MNIST, we used the architecture from \citet{azizzadenesheli2018regularized}, and for our experiments on CIFAR10 and CIFAR100, we trained ten different models, each with a different random seed, using the code from \citet{2017arXiv170508500G}. For MNIST, CIFAR10, and CIFAR100, 10K examples of the training set were reserved as a held-out validation set. Dirichlet shift was simulated on the testing set by sampling with replacement in accordance with class proportions generated by a dirichlet distribution with uniform $\alpha$ values of $0.1$, $1.0$ and $10.0$ (smaller values of $\alpha$ result in more extreme label shift). Samples from the validation set were used for calibration, EM initialization and BBSL \& RLLS confusion matrix estimation. Accuracy was reported on the label-shifted testing set, while the calibration metrics of NLL and ECE (with 15 bins) were reported on the unshifted testing set. In addition to exploring different degrees of dirichlet shift, we also investigated how the algorithms behaved when the number of samples used in the validation and testing set were varied. For example, in experiments with $n=8000$, only 8000 samples from the validation set and 8000 samples from the shifted testing set were presented to the domain adaptation and calibration algorithms. For each model, for a given $\alpha$ and $n$, 10 trials were performed, where each trial consisted of a different sampling (without replacement) of the validation set as well as a different sampling of the dirichlet prior and the label-shifted testing set. This resulted in a total of 100 experiments (10 for each of the 10 different models). Statistical significance was calculated using a signed Wilcoxon test with a one-sided p-value threshold of $0.01$. For MNIST and CIFAR10, we also explored ``tweak one'' shift \citep{bbse}, where the prior of the fourth class was set to a parameter $\rho$ and the remaining class priors were set to $(1-\rho)/9$. We explored $\rho=0.01$ and $\rho=0.9$.

The Kaggle Diabetic Retinopathy dataset \citep{Diabetic40:online} is a collection of retinal fundus images and an associated ``grade'' from 0-4, where 0 indicates healthy and 1-4 indicate progressively more severe stages of retinopathy. For our experiments, we used the publicly-available pretrained model from \citet{JeffreyD27:online}, but modified it so as to make predictions on only one eye at a time (specifically, we supplied the mirror image of a given eye as the input for the second eye). Because test-set labels are unavailable, we separated the validation set used during the training of the model (consisting of 3514 examples) into ``pseudo-validation'' and ``pseudo-test'' sets. Specifically, for each of 100 trials, we sampled $n$ examples from the original validation set without replacement to form a pseudo-validation set, and kept the remaining examples as the pseudo-test set. Calibration was performed on the pseudo-validation set, and calibration metrics of NLL and ECE were reported on the pseudo-test set. Domain shift was then simulated by sampling from the pseudo-test set in such a way that the proportion of ``healthy'' labels was set to a fraction $\rho$, and the relative proportions of retinopathy grades was kept the same as in the source distribution. In the source distribution, $\rho=0.73$; for the simulated domain shift, we explored $\rho=0.5$ and $\rho=0.9$.

\subsection{Max Likelihood With Appropriate Calibration Performs Well At Estimating Shift Weights}

We compared the performance of maximum likelihood (EM), BBSL and RLLS in the presence of different types of calibration, using both MSE of the shift weights as the metric (\textbf{Sec. ~\ref{sec:metrics_for_labelshift}}). Results are in \textbf{Tables 
\ref{tab:cifar10_emvsbbsevsrlls_mseweights_even_dirichlet},
\ref{tab:MNIST_emvsbbsevsrlls_mseweights_even_tweakone},
\ref{tab:cifar100_emvsbbsevsrlls_mseweights_even_dirichlet},
\ref{tab:kaggledr_emvsbbsevsrlls_mseweights_even},
\ref{tab:cifar10_emvsbbsevsrlls_mseweights_even_tweakone} \&
\ref{tab:MNIST_emvsbbsevsrlls_mseweights_even_dirichlet}}. Across all datasets, we observed the following general trends: first, in the absence of calibration, BBSL and RLLS tend to outperform EM, with RLLS tending to perform the best (consistent with the results in \citet{azizzadenesheli2018regularized}). However, as calibration improves, so does the performance of EM. In particular, the best overall performance is achieved when using the variants of temperature scaling that contain class-specific bias parameters - namely BCTS and VS - in combination with EM. 

We also computed the improvement in accuracy achieved by EM with different calibration methods compared to an unadapted baseline (\textbf{Tables ~\ref{tab:cifar10_em_deltaacc_dirichletshift}, ~\ref{tab:cifar100_em_deltaacc_dirichletshift},
~\ref{tab:kaggledr_em_deltaacc},
~\ref{tab:cifar10_em_deltaacc_tweakoneshift}}). Across datasets, we observed that either BCTS or VS tended to achieve the best accuracy. To reconcile this with the observation in \citet{Guo2017-wk} that VS did not give the best ECE compared to TS, we calculated the Negative Log Likelihood (NLL) of different calibration methods on an unshifted test set and found that BCTS and VS tended to achieve the best NLL, even when they did not yield the best ECE (Sec. ~\ref{sec:calibrationcomparison}), indicating that the ECE and NLL metrics do not always agree with each other. Empirically, we found that the NLL corresponds better with the improvement that a calibration method will give to domain adaptation (Sec. ~\ref{sec:nll_corresponds_benefits_labelshift}). This is consistent with other reports stating that ECE computed using only information about the most confidently predicted class, as was done in \citet{Guo2017-wk}, is perhaps not the best metric \citep{vaicenavicius2019evaluating}. 

\section{Discussion}

In this work, we explored the effect of calibration on procedures designed to perform domain adaptation to label shift. In experiments on CIFAR10, MNIST, CIFAR100 and diabetic retinopathy detection, we found that maximum likelihood methods such as EM with specific types of calibration tends to outperform moment matching methods such as BBSL and RLLS. In particular, we found that the best results were achieved when the calibration method involves class-specific bias parameters that can reduce systematic bias in the class probabilities. This capacity for bias correction is absent from the popular Temperature Scaling approach recommended by \citet{Guo2017-wk}. We reconcile this by noting that Guo et al. evaluated calibration using ECE computed on only the most confidently predicted classes, which is known to be misleading \citep{vaicenavicius2019evaluating}, and by observing that Vector Scaling (which does include class-specific bias parameters) performed almost as well as Temperature Scaling in their evaluation.

In addition, we observe that when the calibrated probabilities retain systematic bias, domain adaptation via maximum likelihood is sensitive to the strategy used to compute the source-domain priors. If the source-domain priors $\hat{p}(y=i)$ are not defined in a way that mirrors the systematic bias in the predicted probabilities $\hat{p}(y=i|x)$, then maximum likelihood will estimate a label shift even if the target domain is identical to the source domain (\textbf{Lemma A}) and can produce highly detrimental results (\textbf{Tables ~\ref{tab:kaggledr_diffemsourcepriors_deltaperf}}). By contrast, if the source domain priors for maximum likelihood are specified as recommend in \textbf{Sec. ~\ref{sec:em_sourcedomain_priors_definition}}, maximum likelihood becomes substantially more tolerant of systematic bias in the calibrated probabilities, although it does not tend to outperform BBSL or RLLS in the presence of poor calibration. We conjecture that maximum likelihood is sensitive to systematic bias because the estimate of $\hat{q}(y=i|\boldsymbol{x}_k)$ (e.g. as computed in the E-step of EM) relies heavily on the ratio $\frac{\hat{q}(y=i)}{\hat{p}(y=i)}$. Systematic bias is defined as error in $\hat{p}(y=i)$, which, as it appears in the denominator, could manifest as large errors in $\frac{\hat{q}(y=i)}{\hat{p}(y=i)}$ - particularly when $\hat{p}(y=i)$ is small.

One concern when using EM to find the maximum likelihood estimate is the possibility of getting trapped in local minima. To address this concern, we analyzed the likelihood function and determined that it is concave and bounded (\textbf{Sec. ~\ref{sec:emunimodal}}). Thus, EM (and any standard convex optimization algorithm) converges to the global maximum of the likelihood.

Following the appearance of this work online, an independent study by \citet{unified} verified that our proposed approach of coupling bias-corrected temperature scaling (BCTS) with maximum-likelihood estimation ``uniformly dominates'' compared to BBSL and RLLS. The same study also developed a theoretical analysis of the maximum likelihood approach that confirms the importance of calibration in achieving good performance.

In conclusion, we presented an algorithm that is simple, computationally efficient, and avoids both hyperparameter tuning and the pitfalls associated with retraining deep learning models with importance weighting \citep{importanceweighting}. When tested empirically on a variety of datasets and data shifts, it produces better or comparable results compared to the current state-of-the-art. We posit that maximum likelihood with bias-corrected calibration will prove particularly useful in big data settings where deep learning models are more likely to be deployed.

\section{Author Contributions}
AS conceived of the method and implemented the calibration \& label shift adaptation algorithms. AS and AA designed \& conducted experiments, and wrote the manuscript, with guidance and feedback from AK.

\section{Acknowledgements}

We thank the members of the Kundaje lab for discussion and feedback.

\clearpage

\bibliography{example_paper}
\bibliographystyle{plainnat}

\newpage

\begin{appendix}

\counterwithin{table}{section}
\counterwithin{figure}{section}

\onecolumn


\FloatBarrier
\section{Comparison of Strategies for Initializing EM Source Probabilities}
\FloatBarrier

\begin{table*}[!htbp]
\adjustbox{max width=\textwidth}{
  \centering
  \begin{tabular}{ c | c | c c c | c c c}
    \multirow{2}{*}{\begin{tabular}{c}\textbf{Shift} \\ \textbf{Estimator} \end{tabular}} & \multirow{2}{*}{\begin{tabular}{c}\textbf{Calibration} \\ \textbf{Method} \end{tabular}} & \multicolumn{3}{| c}{$\rho=0.5$} & \multicolumn{3}{| c}{$\rho=0.9$}\\ 
    \cline{3-8}
    & & $n$=500 & $n$=1000 & $n$=1500 & $n$=500 & $n$=1000 & $n$=1500\\
    \hline
    \hline
    EM: source priors from preds & None & \textbf{2.0; 0.0} & \textbf{2.0; 0.0} & \textbf{2.2; 0.0} & \textbf{1.4; 0.0} & \textbf{1.5; 0.0} & \textbf{1.567; 0.0}\\
    EM: source priors from labels & None & -3.4; 1.0 & -3.6; 1.0 & -3.367; 1.0 & 1.0; 1.0 & 1.0; 1.0 & 1.067; 1.0\\
    \hline
    \hline
    EM: source priors from preds & TS & \textbf{2.2; 0.0} & \textbf{2.3; 0.0} & \textbf{2.667; 0.0} & \textbf{1.6; 0.0} & \textbf{2.0; 0.0} & \textbf{2.133; 0.0}\\
    EM: source priors from labels & TS & -64.5; 1.0 & -65.1; 1.0 & -65.267; 1.0 & -91.4; 1.0 & -91.9; 1.0 & -91.833; 1.0\\
    \hline
    \hline
    EM: source priors from preds & NBVS & \textbf{3.5; 0.0} & \textbf{4.1; 0.0} & \textbf{4.233; 0.0} & \textbf{2.2; 0.0} & \textbf{2.4; 0.0} & \textbf{2.467; 0.0}\\
    EM: source priors from labels & NBVS & -5.6; 1.0 & -4.6; 1.0 & -4.567; 1.0 & 0.4; 1.0 & 0.6; 1.0 & 0.7; 1.0\\
    \hline
    \hline
    EM: source priors from preds & BCTS & \textbf{3.8; 0.0} & 4.65; 0.0 & 4.633; 0.0 & 3.6; 0.0 & 3.6; 0.0 & 3.733; 0.0\\
    EM: source priors from labels & BCTS & 3.8; 1.0 & 4.65; 1.0 & 4.633; 1.0 & 3.6; 1.0 & 3.6; 1.0 & 3.733; 1.0\\
    \hline
    \hline
    EM: source priors from preds & VS & 3.8; 0.0 & 4.4; 0.0 & 4.633; 0.0 & 3.5; 0.0 & 3.6; 0.0 & 3.733; 0.0\\
    EM: source priors from labels & VS & 3.8; 1.0 & 4.4; 1.0 & 4.633; 1.0 & 3.5; 1.0 & 3.6; 1.0 & 3.733; 1.0\\
  \end{tabular}}
  \caption{\textbf{The strategy for computing EM source priors heavily affects domain adaptation if probabilities retain systematic bias}. Value before the semicolon is the median improvement in \%accuracy (across 100 trials) caused by applying domain adaptation to the predictions on a diabetic retinopathy prediction task. Value after the semicolon is the median rank of a particular method relative to the other method in the pair. Domain shift is induced by varying the proportion of ``healthy'' examples $\rho$; in the source distribution, $\rho=0.73$. We see that calibration methods that lack class-specific bias parameters (i.e. no calibration, TS and NBVS) can hurt domain adaptation if source priors are initialized by averaging true labels rather than the predicted probabilities. A bold value in a pair is significantly better than the non-bold value according to a paired Wilcoxon test at $p \le 0.01$. See \textbf{Sec. ~\ref{sec:experimentalsetup}} for details on the experimental setup.}
  \label{tab:kaggledr_diffemsourcepriors_deltaperf}
\end{table*}

\FloatBarrier
\section{Calibration Quality Comparison}
\label{sec:calibrationcomparison}
\FloatBarrier

We find that bias-corrected versions of Temperature Scaling (namely Bias-Corrected Temperature Scaling and Vector Scaling) tend to yield the best Negative Log Likelihood on an unshifted test set, even if they do not always yield the best ECE. Results are shown in the tables below.

\begin{table*}
\adjustbox{max width=\textwidth}{
  \centering
  \begin{tabular}{ c | c c c | c c c }
    \multirow{2}{*}{\begin{tabular}{c}\textbf{Calibration} \\ \textbf{Method} \end{tabular}} & \multicolumn{3}{| c}{NLL} & \multicolumn{3}{| c}{ECE}\\
    \cline{2-7}
    & $n$=2000 & $n$=4000 & $n$=8000 & $n$=2000 & $n$=4000 & $n$=8000\\
    \hline
    None & 0.299; 4.0 & 0.299; 4.0 & 0.299; 4.0 & 2.696; 4.0 & 2.696; 4.0 & 2.696; 4.0\\
    TS & 0.292; 3.0 & 0.292; 3.0 & 0.292; 3.0 & \textbf{1.074; 1.0} & 1.029; 2.0 & 0.996; 3.0\\
    NBVS & 0.277; 2.0 & 0.276; 2.0 & 0.276; 2.0 & 1.09; 1.5 & 0.997; 1.5 & \textbf{0.932; 1.5}\\
    BCTS & \textbf{0.274; 0.0} & \textbf{0.273; 1.0} & 0.272; 1.0 & \textbf{1.046; 1.0} & \textbf{0.963; 1.0} & \textbf{0.921; 1.0}\\
    VS & 0.276; 1.0 & \textbf{0.274; 0.5} & \textbf{0.272; 0.0} & 1.175; 2.5 & 1.022; 2.0 & 0.966; 1.0\\
  \end{tabular}}
  \caption{\textbf{CIFAR10: NLL and ECE for different calibration methods}. Metrics were computed on a test set that had the same distribution as the validation set. Value before the semicolon is the median of the metric over all the runs. Value after the semicolon is the median rank of the method relative to other methods in the column. $n$ indicates the number of examples used for calibratin. Bold values in a column are not significantly different from the best performing method in the column, as measured by a paired Wilcoxon test at $p \le 0.01$. See \textbf{Sec. ~\ref{sec:experimentalsetup}} for details on the experimental setup.}
  \label{tab:cifar10_calibrationcomparison}
\end{table*}

\begin{table*}
\adjustbox{max width=\textwidth}{
  \centering
  \begin{tabular}{ c | c c c | c c c }
    \multirow{2}{*}{\begin{tabular}{c}\textbf{Calibration} \\ \textbf{Method} \end{tabular}} & \multicolumn{3}{| c}{NLL} & \multicolumn{3}{| c}{ECE}\\
    \cline{2-7}
    & $n$=7000 & $n$=8500 & $n$=10000 & $n$=7000 & $n$=8500 & $n$=10000\\
    \hline
    None & 1.727; 4.0 & 1.727; 4.0 & 1.727; 4.0 & 19.999; 4.0 & 19.999; 4.0 & 19.999; 4.0\\
    TS & 1.281; 3.0 & 1.281; 3.0 & 1.281; 3.0 & 3.149; 3.0 & 3.156; 3.0 & 3.184; 3.0\\
    NBVS & 1.236; 2.0 & 1.235; 2.0 & 1.234; 2.0 & \textbf{2.291; 0.0} & \textbf{2.289; 0.0} & \textbf{2.344; 0.0}\\
    BCTS & 1.23; 1.0 & 1.228; 1.0 & 1.227; 1.0 & 2.91; 2.0 & 2.943; 2.0 & 2.918; 2.0\\
    VS & \textbf{1.229; 0.0} & \textbf{1.225; 0.0} & \textbf{1.224; 0.0} & 2.495; 1.0 & 2.501; 1.0 & 2.474; 1.0\\
  \end{tabular}}
  \caption{\textbf{CIFAR100: NLL and ECE for different calibration methods}. Analogous to ~\textbf{Table ~\ref{tab:cifar10_calibrationcomparison}}.}
  \label{tab:cifar100_calibrationcomparison}
\end{table*}

\begin{table*}
\adjustbox{max width=\textwidth}{
  \centering
  \begin{tabular}{ c | c c c | c c c }
    \multirow{2}{*}{\begin{tabular}{c}\textbf{Calibration} \\ \textbf{Method} \end{tabular}} & \multicolumn{3}{| c}{NLL} & \multicolumn{3}{| c}{ECE}\\
    \cline{2-7}
    & $n$=500 & $n$=1000 & $n$=1500 & $n$=500 & $n$=1000 & $n$=1500\\
    \hline
    None & 0.641; 4.0 & 0.64; 4.0 & 0.638; 4.0 & 8.738; 4.0 & 8.746; 4.0 & 8.745; 4.0\\
    TS & 0.57; 3.0 & 0.57; 3.0 & 0.568; 3.0 & 3.636; 3.0 & 3.749; 3.0 & 3.872; 3.0\\
    NBVS & 0.542; 2.0 & 0.54; 2.0 & 0.539; 2.0 & \textbf{1.956; 0.0} & \textbf{2.012; 1.0} & \textbf{2.047; 1.0}\\
    BCTS & \textbf{0.513; 0.0} & \textbf{0.511; 1.0} & 0.509; 1.0 & 2.157; 1.0 & \textbf{2.135; 1.0} & \textbf{2.08; 1.0}\\
    VS & 0.518; 1.0 & \textbf{0.512; 0.0} & 0.509; 0.0 & 2.187; 1.0 & \textbf{2.006; 1.0} & \textbf{2.014; 1.0}\\
  \end{tabular}}
  \caption{\textbf{Kaggle Diabetic Retinopathy Detection: NLL and ECE for different calibration methods}. Analogous to ~\textbf{Table ~\ref{tab:cifar10_calibrationcomparison}}.}
  \label{tab:kaggledr_calibrationcomparison}
\end{table*}

\FloatBarrier
\section{CIFAR10 Supplementary Tables}
\FloatBarrier

\begin{table*}[!htbp]
\adjustbox{max width=\textwidth}{
  \centering
  \begin{tabular}{ c | c | c c c | c c c}
    \multirow{2}{*}{\begin{tabular}{c}\textbf{Shift} \\ \textbf{Estimator} \end{tabular}} & \multirow{2}{*}{\begin{tabular}{c}\textbf{Calibration} \\ \textbf{Method} \end{tabular}} & \multicolumn{3}{| c}{$\rho=0.01$} & \multicolumn{3}{| c}{$\rho=0.9$}\\ 
    \cline{3-8}
    & & $n$=2000 & $n$=4000 & $n$=8000 & $n$=2000 & $n$=4000 & $n$=8000\\
    \hline
    \hline
    EM & None & 0.75; 3.0 & 0.825; 3.0 & 0.813; 3.0 & 16.025; 4.0 & 15.837; 4.0 & 15.944; 4.0\\
    EM & TS & 0.775; 3.0 & 0.812; 3.0 & 0.862; 3.0 & 16.675; 3.0 & 16.412; 3.0 & 16.45; 3.0\\
    EM & NBVS & \textbf{1.1; 1.0} & 1.162; 2.0 & 1.169; 2.0 & 17.1; 2.0 & 17.062; 2.0 & 17.244; 2.0\\
    EM & BCTS & \textbf{1.2; 1.0} & 1.25; 1.0 & 1.194; 1.0 & \textbf{17.3; 1.0} & 17.025; 1.0 & 17.362; 1.0\\
    EM & VS & \textbf{1.2; 1.0} & \textbf{1.275; 1.0} & \textbf{1.212; 1.0} & \textbf{17.125; 1.0} & \textbf{17.225; 1.0} & \textbf{17.419; 0.0}\\
  \end{tabular}}
  \caption{\textbf{CIFAR10: Comparison of calibration methods when using EM adaptation to ``tweak-one'' shift, with $\Delta$\%accuracy as the metric.} Analogous to \textbf{Table ~\ref{tab:cifar10_em_deltaacc_dirichletshift}}.}
  \label{tab:cifar10_em_deltaacc_tweakoneshift}
\end{table*}

\begin{table*}[!htbp]
\adjustbox{max width=\textwidth}{
  \centering
  \begin{tabular}{ c | c | c c c | c c c | c c c}
    \multirow{2}{*}{\begin{tabular}{c}\textbf{Shift} \\ \textbf{Estimator} \end{tabular}} & \multirow{2}{*}{\begin{tabular}{c}\textbf{Calibration} \\ \textbf{Method} \end{tabular}} & \multicolumn{3}{| c}{$\alpha=0.1$} & \multicolumn{3}{| c}{$\alpha=1.0$} & \multicolumn{3}{| c}{$\alpha=10$}\\ 
    \cline{3-11}
    & & $n$=2000 & $n$=4000 & $n$=8000 & $n$=2000 & $n$=4000 & $n$=8000 & $n$=2000 & $n$=4000 & $n$=8000\\
    \hline
    \hline
    EM & None & 0.01049; 15.0 & 0.00763; 15.0 & 0.00795; 15.0 & 0.00283; 15.0 & 0.00202; 15.0 & 0.00147; 15.0 & 0.00191; 12.0 & 0.001; 10.0 & 0.0006; 14.0\\
    EM & TS & 0.00944; 15.0 & 0.00705; 15.0 & 0.00693; 15.0 & 0.00262; 14.0 & 0.00192; 15.0 & 0.00148; 15.0 & 0.00195; 13.0 & 0.00101; 13.0 & 0.00058; 14.0\\
    EM & NBVS & 0.0014; 2.0 & 0.00106; 2.0 & 0.00075; 4.0 & 0.00133; 2.0 & 0.00071; 2.0 & 0.00043; 3.0 & \textbf{0.00161; 6.0} & \textbf{0.00079; 4.5} & \textbf{0.00047; 6.0}\\
    EM & BCTS & \textbf{0.00037; 1.0} & \textbf{0.00034; 1.0} & \textbf{0.00022; 1.0} & \textbf{0.00126; 1.0} & 0.00071; 2.0 & \textbf{0.00043; 2.0} & \textbf{0.00161; 6.0} & \textbf{0.00077; 4.0} & \textbf{0.00047; 5.5}\\
    EM & VS & 0.00061; 1.0 & \textbf{0.00031; 1.0} & \textbf{0.0002; 1.0} & \textbf{0.00118; 1.0} & \textbf{0.00067; 2.0} & \textbf{0.0004; 1.5} & \textbf{0.00167; 6.0} & \textbf{0.00076; 4.5} & \textbf{0.00045; 6.0}\\
    BBSL-hard & None & 0.00443; 13.0 & 0.00214; 13.0 & 0.00125; 13.0 & 0.00266; 14.0 & 0.00135; 14.0 & 0.00074; 13.0 & 0.00244; 15.0 & 0.00123; 15.0 & 0.00055; 13.0\\
    BBSL-soft & None & 0.00309; 9.0 & 0.00133; 9.0 & 0.00092; 9.0 & 0.00188; 9.0 & 0.00108; 9.0 & 0.00057; 7.5 & 0.00187; 9.0 & 0.0009; 9.0 & 0.00047; 9.0\\
    BBSL-soft & TS & 0.0031; 8.0 & 0.00132; 7.0 & 0.00091; 8.0 & 0.00191; 10.0 & 0.00102; 8.0 & 0.0006; 8.0 & 0.00179; 7.0 & 0.00095; 10.0 & 0.00048; 9.0\\
    BBSL-soft & NBVS & 0.0028; 9.0 & 0.00137; 8.0 & 0.00098; 8.0 & 0.00175; 8.0 & 0.00096; 7.0 & 0.00061; 7.0 & 0.00179; 7.0 & 0.00081; 7.0 & 0.00048; 7.0\\
    BBSL-soft & BCTS & 0.00288; 9.0 & 0.00126; 7.0 & 0.00104; 8.0 & 0.00171; 8.0 & 0.00102; 8.0 & 0.0006; 7.0 & 0.00176; 7.0 & 0.00083; 8.0 & 0.00048; 7.0\\
    BBSL-soft & VS & 0.00306; 9.0 & 0.00134; 9.0 & 0.00103; 8.0 & 0.00173; 8.0 & 0.00102; 7.0 & 0.00061; 7.0 & 0.00172; 7.0 & 0.00082; 7.0 & 0.00047; 7.0\\
    RLLS-hard & None & 0.00405; 13.0 & 0.00209; 13.0 & 0.00124; 12.0 & 0.00265; 13.0 & 0.00135; 13.0 & 0.00074; 13.0 & 0.00244; 15.0 & 0.00123; 15.0 & 0.00055; 13.0\\
    RLLS-soft & None & 0.00316; 9.0 & 0.00127; 8.5 & 0.0009; 8.0 & 0.00183; 10.0 & 0.00108; 10.0 & 0.00057; 8.0 & 0.00187; 8.5 & 0.0009; 9.0 & 0.00047; 9.0\\
    RLLS-soft & TS & 0.00312; 8.0 & 0.00125; 7.0 & 0.0009; 7.0 & 0.00186; 9.0 & 0.00102; 8.0 & 0.0006; 8.0 & 0.00179; 7.0 & 0.00095; 9.0 & 0.00048; 9.0\\
    RLLS-soft & NBVS & 0.00275; 8.0 & 0.00131; 8.0 & 0.00094; 7.5 & 0.00168; 7.0 & 0.00096; 7.0 & 0.00061; 7.5 & 0.00179; 7.0 & 0.00081; 7.0 & 0.00048; 7.0\\
    RLLS-soft & BCTS & 0.00284; 7.0 & 0.00124; 7.0 & 0.00093; 7.0 & 0.00169; 8.0 & 0.00102; 8.0 & 0.0006; 7.0 & 0.00176; 7.0 & 0.00083; 8.0 & 0.00048; 7.0\\
    RLLS-soft & VS & 0.00298; 7.5 & 0.00133; 8.0 & 0.00091; 8.0 & 0.00171; 7.0 & 0.00102; 7.0 & 0.00061; 7.0 & 0.00172; 7.0 & 0.00082; 7.0 & 0.00047; 7.0\\
  \end{tabular}}
  \caption{\textbf{CIFAR10: Comparison of all calibration and domain adaptation methods, using MSE (Sec. ~\ref{sec:metrics_for_labelshift}) as the metric (dirichlet shift).} Value before the semicolon is the median of the metric over all trials. Value after the semicolon is the median rank of the domain adaptation + calibration method combination relative to the other method combinations in the column. Bold values in a column are not significantly different from the best-performing method in the column as measured by a paired Wilcoxon test at $p < 0.01$. EM with BCTS or VS tends to achieve the best performance. See \textbf{Sec. ~\ref{sec:experimentalsetup}} for details on the experimental setup.}
  \label{tab:cifar10_all_mseweights_even_dirichletshift}
\end{table*}

\begin{table*}[!htbp]
\adjustbox{max width=\textwidth}{
  \centering
  \begin{tabular}{ c | c | c c c | c c c}
    \multirow{2}{*}{\begin{tabular}{c}\textbf{Shift} \\ \textbf{Estimator} \end{tabular}} & \multirow{2}{*}{\begin{tabular}{c}\textbf{Calibration} \\ \textbf{Method} \end{tabular}} & \multicolumn{3}{| c}{$\rho=0.01$} & \multicolumn{3}{| c}{$\rho=0.9$}\\ 
    \cline{3-8}
    & & $n$=2000 & $n$=4000 & $n$=8000 & $n$=2000 & $n$=4000 & $n$=8000\\
    \hline
    \hline
    EM & None & 0.00177; 13.0 & 0.00094; 13.0 & 0.00066; 15.0 & 0.04165; 15.0 & 0.03988; 15.0 & 0.03791; 15.0\\
    EM & TS & 0.00183; 13.0 & 0.00101; 14.0 & 0.00071; 15.0 & 0.04156; 16.0 & 0.04897; 16.0 & 0.04237; 16.0\\
    EM & NBVS & \textbf{0.00139; 3.0} & \textbf{0.0006; 3.0} & 0.0004; 4.0 & 0.00274; 2.0 & 0.00187; 2.0 & 0.00172; 2.0\\
    EM & BCTS & \textbf{0.00132; 3.0} & \textbf{0.00058; 2.0} & \textbf{0.00036; 2.0} & \textbf{0.00125; 1.0} & \textbf{0.00083; 1.0} & \textbf{0.00072; 1.0}\\
    EM & VS & \textbf{0.00139; 3.5} & \textbf{0.00057; 2.0} & \textbf{0.00038; 3.5} & \textbf{0.00165; 1.0} & \textbf{0.00083; 1.0} & \textbf{0.00066; 1.0}\\
    BBSL-hard & None & 0.00228; 15.0 & 0.001; 15.0 & 0.00054; 14.0 & 0.01524; 13.0 & 0.00834; 13.0 & 0.00473; 10.0\\
    BBSL-soft & None & 0.00148; 9.0 & 0.00067; 8.0 & 0.0004; 6.0 & 0.01356; 11.0 & 0.00779; 10.0 & 0.00488; 10.0\\
    BBSL-soft & TS & 0.00146; 7.0 & 0.00068; 8.0 & 0.0004; 6.0 & 0.01411; 10.0 & 0.00852; 11.0 & 0.00565; 12.0\\
    BBSL-soft & NBVS & 0.00153; 8.0 & 0.00073; 7.0 & 0.00043; 8.0 & 0.01187; 7.5 & 0.00636; 7.0 & 0.00346; 7.0\\
    BBSL-soft & BCTS & 0.00152; 8.0 & 0.00073; 8.0 & 0.00042; 7.0 & 0.01189; 8.0 & 0.00607; 7.0 & 0.00351; 6.0\\
    BBSL-soft & VS & 0.00159; 8.0 & 0.00075; 7.5 & 0.00043; 9.5 & 0.0121; 8.0 & 0.00607; 6.0 & 0.0033; 5.0\\
    RLLS-hard & None & 0.00228; 15.0 & 0.001; 14.0 & 0.00054; 13.0 & 0.01347; 13.0 & 0.00815; 13.0 & 0.00476; 10.0\\
    RLLS-soft & None & 0.00148; 9.0 & 0.00067; 8.0 & 0.0004; 6.0 & 0.01355; 10.0 & 0.00779; 11.0 & 0.00488; 11.0\\
    RLLS-soft & TS & 0.00146; 7.5 & 0.00068; 7.0 & 0.0004; 5.0 & 0.01368; 10.0 & 0.00832; 12.0 & 0.00565; 13.0\\
    RLLS-soft & NBVS & 0.00153; 8.0 & 0.00073; 7.0 & 0.00043; 7.0 & 0.01155; 8.0 & 0.00626; 7.0 & 0.00343; 8.0\\
    RLLS-soft & BCTS & 0.00152; 7.0 & 0.00073; 8.0 & 0.00042; 7.0 & 0.01178; 7.0 & 0.00601; 7.0 & 0.00337; 7.0\\
    RLLS-soft & VS & 0.00159; 7.0 & 0.00075; 7.0 & 0.00043; 9.0 & 0.01218; 7.0 & 0.00598; 6.0 & 0.00326; 6.0\\
  \end{tabular}}
  \caption{\textbf{CIFAR10: Comparison of all calibration and domain adaptation methods, using MSE (Sec. ~\ref{sec:metrics_for_labelshift}) as the metric (``tweak-one'' shift).} Value before the semicolon is the median of the metric over all trials. Value after the semicolon is the median rank of the domain adaptation + calibration method combination relative to the other method combinations in the column. Bold values in a column are not significantly different from the best-performing method in the column as measured by a paired Wilcoxon test at $p < 0.01$. EM with BCTS or VS tends to achieve the best performance. See \textbf{Sec. ~\ref{sec:experimentalsetup}} for details on the experimental setup.}
  \label{tab:cifar10_all_mseweights_even_tweakoneshift}
\end{table*}

\begin{table*}[!htbp]
\adjustbox{max width=\textwidth}{
  \centering
  \begin{tabular}{ c | c | c c c | c c c}
    \multirow{2}{*}{\begin{tabular}{c}\textbf{Shift} \\ \textbf{Estimator} \end{tabular}} & \multirow{2}{*}{\begin{tabular}{c}\textbf{Calibration} \\ \textbf{Method} \end{tabular}} & \multicolumn{3}{| c}{$\rho=0.01$} & \multicolumn{3}{| c}{$\rho=0.9$}\\ 
    \cline{3-8}
    & & $n$=2000 & $n$=4000 & $n$=8000 & $n$=2000 & $n$=4000 & $n$=8000\\
    \hline
    \hline
    EM & None & 0.00177; 2.0 & 0.00094; 2.0 & 0.00066; 4.0 & 0.04165; 4.0 & 0.03988; 4.0 & 0.03791; 4.0\\
    BBSL-hard & None & 0.00228; 3.0 & 0.001; 3.0 & 0.00054; 3.0 & \textbf{0.01524; 2.0} & 0.00834; 2.0 & \textbf{0.00473; 1.5}\\
    BBSL-soft & None & 0.00148; 1.0 & 0.00067; 1.0 & 0.0004; 1.0 & \textbf{0.01356; 1.0} & \textbf{0.00779; 1.0} & \textbf{0.00488; 2.0}\\
    RLLS-hard & None & 0.00228; 3.0 & 0.001; 3.0 & 0.00054; 3.0 & \textbf{0.01347; 2.0} & 0.00815; 2.0 & \textbf{0.00476; 1.0}\\
    RLLS-soft & None & \textbf{0.00148; 1.0} & \textbf{0.00067; 1.0} & 0.0004; 1.0 & \textbf{0.01355; 1.0} & 0.00779; 1.0 & 0.00488; 2.0\\
    \hline
    \hline
    EM & TS & 0.00183; 2.0 & 0.00101; 2.0 & 0.00071; 2.0 & 0.04156; 2.0 & 0.04897; 2.0 & 0.04237; 2.0\\
    BBSL-soft & TS & 0.00146; 1.0 & 0.00068; 1.0 & 0.0004; 1.0 & \textbf{0.01411; 1.0} & 0.00852; 0.0 & \textbf{0.00565; 0.0}\\
    RLLS-soft & TS & \textbf{0.00146; 1.0} & \textbf{0.00068; 0.0} & \textbf{0.0004; 0.0} & \textbf{0.01368; 1.0} & 0.00832; 1.0 & 0.00565; 1.0\\
    \hline
    \hline
    EM & NBVS & \textbf{0.00139; 0.0} & \textbf{0.0006; 0.0} & \textbf{0.0004; 0.0} & \textbf{0.00274; 0.0} & \textbf{0.00187; 0.0} & \textbf{0.00172; 0.0}\\
    BBSL-soft & NBVS & 0.00153; 2.0 & 0.00073; 1.0 & 0.00043; 1.0 & 0.01187; 1.0 & 0.00636; 1.0 & 0.00346; 1.0\\
    RLLS-soft & NBVS & 0.00153; 1.0 & 0.00073; 1.0 & 0.00043; 1.0 & 0.01155; 2.0 & 0.00626; 2.0 & 0.00343; 2.0\\
    \hline
    \hline
    EM & BCTS & \textbf{0.00132; 0.0} & \textbf{0.00058; 0.0} & \textbf{0.00036; 0.0} & \textbf{0.00125; 0.0} & \textbf{0.00083; 0.0} & \textbf{0.00072; 0.0}\\
    BBSL-soft & BCTS & 0.00152; 1.5 & 0.00073; 1.0 & 0.00042; 2.0 & 0.01189; 1.0 & 0.00607; 1.0 & 0.00351; 1.0\\
    RLLS-soft & BCTS & 0.00152; 1.0 & 0.00073; 1.0 & 0.00042; 1.0 & 0.01178; 2.0 & 0.00601; 2.0 & 0.00337; 2.0\\
    \hline
    \hline
    EM & VS & \textbf{0.00139; 0.0} & \textbf{0.00057; 0.0} & \textbf{0.00038; 0.0} & \textbf{0.00165; 0.0} & \textbf{0.00083; 0.0} & \textbf{0.00066; 0.0}\\
    BBSL-soft & VS & 0.00159; 2.0 & 0.00075; 1.0 & 0.00043; 2.0 & 0.0121; 1.0 & 0.00607; 1.0 & 0.0033; 1.0\\
    RLLS-soft & VS & 0.00159; 1.0 & 0.00075; 1.0 & 0.00043; 1.0 & 0.01218; 2.0 & 0.00598; 2.0 & 0.00326; 2.0\\
  \end{tabular}}
  \caption{\textbf{CIFAR10: Comparison of EM, BBSL and RLLS (``tweak-one'' shift) using MSE as the metric.} Analogous to \textbf{Table ~\ref{tab:cifar10_emvsbbsevsrlls_mseweights_even_dirichlet}}, but with tweak-one shift instead of dirichlet shift.}
  \label{tab:cifar10_emvsbbsevsrlls_mseweights_even_tweakone}
\end{table*}

\FloatBarrier
\section{MNIST Tables}
\FloatBarrier

\begin{table*}[!htbp]
\adjustbox{max width=\textwidth}{
  \centering
  \begin{tabular}{ c | c | c c c | c c c | c c c}
    \multirow{2}{*}{\begin{tabular}{c}\textbf{Shift} \\ \textbf{Estimator} \end{tabular}} & \multirow{2}{*}{\begin{tabular}{c}\textbf{Calibration} \\ \textbf{Method} \end{tabular}} & \multicolumn{3}{| c}{$\alpha=0.1$} & \multicolumn{3}{| c}{$\alpha=1.0$} & \multicolumn{3}{| c}{$\alpha=10$}\\ 
    \cline{3-11}
    & & $n$=2000 & $n$=4000 & $n$=8000 & $n$=2000 & $n$=4000 & $n$=8000 & $n$=2000 & $n$=4000 & $n$=8000\\
    \hline
    \hline
    EM & None & 0.00438; 13.0 & 0.00294; 6.0 & 0.00227; 13.0 & 0.00339; 16.0 & 0.00256; 16.0 & 0.00207; 16.0 & 0.00217; 15.0 & 0.00133; 16.0 & 0.00093; 16.0\\
    EM & TS & 0.00322; 5.5 & 0.00195; 4.0 & 0.00189; 9.5 & 0.00203; 5.0 & 0.00102; 5.0 & 0.00084; 10.5 & 0.00158; 7.5 & 0.00081; 7.0 & 0.00052; 8.0\\
    EM & NBVS & 0.00144; 2.0 & 0.00104; 2.0 & 0.00085; 2.0 & \textbf{0.00167; 3.0} & \textbf{0.00094; 2.0} & 0.0007; 3.0 & 0.00165; 8.5 & 0.00081; 6.5 & 0.00051; 6.5\\
    EM & BCTS & \textbf{0.00085; 1.0} & \textbf{0.00057; 1.0} & 0.0004; 1.0 & \textbf{0.00165; 2.0} & \textbf{0.00093; 2.0} & 0.0007; 3.0 & 0.00157; 5.0 & \textbf{0.0008; 4.0} & \textbf{0.00049; 4.0}\\
    EM & VS & \textbf{0.00087; 1.0} & \textbf{0.00056; 1.0} & \textbf{0.00035; 1.0} & \textbf{0.00172; 2.0} & \textbf{0.00095; 2.0} & 0.00073; 3.0 & 0.00156; 8.0 & \textbf{0.0008; 7.0} & 0.00054; 7.0\\
    BBSL-hard & None & 0.00491; 13.0 & 0.0036; 11.0 & 0.00244; 12.5 & 0.00276; 14.0 & 0.00167; 14.0 & 0.00125; 14.0 & 0.00208; 15.0 & 0.00106; 14.0 & 0.00073; 15.0\\
    BBSL-soft & None & 0.00459; 10.0 & 0.00324; 11.0 & 0.00213; 10.5 & 0.00227; 9.0 & 0.00137; 9.0 & 0.00085; 9.0 & 0.00174; 9.0 & 0.00083; 9.0 & 0.00052; 7.5\\
    BBSL-soft & TS & 0.00417; 9.0 & 0.00309; 9.0 & 0.00218; 8.0 & 0.00232; 7.0 & 0.00129; 6.0 & 0.00083; 7.0 & 0.00157; 5.0 & \textbf{0.00079; 6.0} & \textbf{0.00049; 5.0}\\
    BBSL-soft & NBVS & 0.00428; 10.5 & 0.00306; 10.0 & 0.0022; 10.0 & 0.00234; 8.0 & 0.00129; 7.5 & 0.00082; 7.0 & 0.00171; 7.0 & 0.00085; 7.5 & 0.00052; 7.0\\
    BBSL-soft & BCTS & 0.00413; 10.0 & 0.00303; 10.0 & 0.00228; 9.0 & 0.00228; 8.0 & 0.0013; 8.0 & 0.00081; 6.0 & 0.00168; 6.0 & 0.00083; 6.5 & \textbf{0.0005; 6.0}\\
    BBSL-soft & VS & 0.0039; 9.0 & 0.00322; 9.5 & 0.00221; 10.0 & 0.0022; 9.0 & 0.00136; 9.0 & 0.00085; 9.0 & 0.00163; 8.0 & 0.00084; 7.0 & 0.00054; 9.0\\
    RLLS-hard & None & 0.00472; 13.0 & 0.00347; 10.5 & 0.00238; 12.0 & 0.00271; 13.5 & 0.00167; 14.0 & 0.00125; 13.0 & 0.00208; 14.0 & 0.00106; 14.0 & 0.00073; 14.0\\
    RLLS-soft & None & 0.0044; 9.0 & 0.00314; 10.0 & 0.00211; 9.0 & 0.00226; 9.0 & 0.00137; 10.0 & 0.00085; 9.0 & 0.00174; 9.0 & 0.00083; 10.0 & 0.00052; 8.0\\
    RLLS-soft & TS & 0.00407; 8.0 & 0.00289; 8.0 & 0.00206; 8.0 & 0.00232; 7.0 & 0.00129; 7.0 & 0.00083; 7.0 & 0.00157; 5.0 & \textbf{0.00079; 6.0} & \textbf{0.00049; 5.0}\\
    RLLS-soft & NBVS & 0.00409; 9.0 & 0.00284; 9.0 & 0.00211; 9.0 & 0.00232; 9.0 & 0.00129; 8.0 & 0.00082; 7.0 & 0.00171; 7.0 & 0.00085; 8.0 & 0.00052; 8.0\\
    RLLS-soft & BCTS & 0.00399; 8.0 & 0.00283; 8.0 & 0.00213; 8.0 & 0.00226; 7.5 & 0.0013; 8.0 & 0.00082; 6.0 & 0.00168; 6.5 & 0.00083; 6.0 & \textbf{0.0005; 6.0}\\
    RLLS-soft & VS & 0.00385; 8.0 & 0.00282; 8.0 & 0.00212; 9.0 & 0.0022; 8.5 & 0.00136; 9.0 & 0.00085; 9.0 & 0.00163; 8.0 & 0.00084; 8.0 & 0.00054; 9.0\\
  \end{tabular}}
  \caption{\textbf{MNIST: Comparison of all calibration and domain adaptation methods, using MSE (Sec. ~\ref{sec:metrics_for_labelshift}) as the metric (dirichlet shift).} Value before the semicolon is the median of the metric over all trials. Value after the semicolon is the median rank of the domain adaptation + calibration method combination relative to the other method combinations in the column. Bold values in a column are not significantly different from the best-performing method in the column as measured by a paired Wilcoxon test at $p < 0.01$. EM with BCTS or VS tends to achieve the best performance, particularly for larger amounts of shift (corresponding to smaller $\alpha$). See \textbf{Sec. ~\ref{sec:experimentalsetup}} for details on the experimental setup.}
  \label{tab:MNIST_all_mseweights_even_dirichletshift}
\end{table*}

\begin{table*}[!htbp]
\adjustbox{max width=\textwidth}{
  \centering
  \begin{tabular}{ c | c | c c c | c c c}
    \multirow{2}{*}{\begin{tabular}{c}\textbf{Shift} \\ \textbf{Estimator} \end{tabular}} & \multirow{2}{*}{\begin{tabular}{c}\textbf{Calibration} \\ \textbf{Method} \end{tabular}} & \multicolumn{3}{| c}{$\rho=0.01$} & \multicolumn{3}{| c}{$\rho=0.9$}\\ 
    \cline{3-8}
    & & $n$=2000 & $n$=4000 & $n$=8000 & $n$=2000 & $n$=4000 & $n$=8000\\
    \hline
    \hline
    EM & None & 0.00197; 14.0 & 0.00107; 14.0 & 0.00067; 14.0 & 0.00584; 11.0 & 0.00305; 5.0 & 0.00316; 12.0\\
    EM & TS & 0.00162; 7.0 & 0.00084; 5.0 & 0.00054; 3.0 & 0.00318; 3.0 & 0.00155; 3.0 & 0.00085; 3.0\\
    EM & NBVS & 0.00158; 5.0 & 0.00076; 3.5 & 0.00054; 2.0 & 0.00112; 2.0 & 0.00059; 2.0 & 0.00038; 2.0\\
    EM & BCTS & \textbf{0.00153; 3.5} & \textbf{0.00076; 3.5} & \textbf{0.00052; 2.0} & \textbf{0.0006; 1.0} & \textbf{0.00037; 0.0} & \textbf{0.00028; 0.0}\\
    EM & VS & 0.00155; 5.0 & \textbf{0.00076; 3.0} & \textbf{0.00052; 2.0} & 0.00067; 1.0 & 0.00042; 1.0 & 0.00033; 1.0\\
    BBSL-hard & None & 0.00197; 15.0 & 0.00122; 16.0 & 0.00079; 16.0 & 0.00897; 15.0 & 0.00689; 16.0 & 0.00608; 16.0\\
    BBSL-soft & None & 0.00161; 10.0 & 0.00099; 10.0 & 0.00057; 11.0 & 0.00662; 12.0 & 0.00379; 6.0 & 0.00298; 5.0\\
    BBSL-soft & TS & 0.00161; 6.0 & 0.00091; 6.0 & 0.00054; 6.0 & 0.00659; 11.0 & 0.00395; 10.0 & 0.00339; 10.0\\
    BBSL-soft & NBVS & 0.00162; 9.0 & 0.00088; 9.0 & 0.00057; 9.0 & 0.00607; 10.0 & 0.00385; 9.0 & 0.00349; 9.0\\
    BBSL-soft & BCTS & 0.00158; 7.0 & 0.00087; 7.5 & 0.00055; 7.0 & 0.00623; 9.0 & 0.00383; 9.0 & 0.00338; 8.0\\
    BBSL-soft & VS & 0.00171; 9.5 & 0.00086; 9.0 & 0.00056; 11.0 & 0.00641; 10.0 & 0.00407; 11.0 & 0.00393; 13.0\\
    RLLS-hard & None & 0.00197; 15.0 & 0.00122; 15.0 & 0.00079; 15.0 & 0.00868; 14.0 & 0.00676; 15.0 & 0.00593; 15.0\\
    RLLS-soft & None & 0.00161; 9.0 & 0.00099; 9.0 & 0.00057; 10.0 & 0.00622; 10.0 & 0.00379; 6.0 & 0.00298; 5.0\\
    RLLS-soft & TS & 0.00161; 6.0 & 0.00091; 6.0 & 0.00054; 5.0 & 0.00627; 9.0 & 0.00393; 9.0 & 0.00339; 10.0\\
    RLLS-soft & NBVS & 0.00162; 8.0 & 0.00088; 8.0 & 0.00057; 8.0 & 0.006; 7.0 & 0.0038; 9.0 & 0.00349; 9.0\\
    RLLS-soft & BCTS & 0.00158; 6.0 & 0.00087; 7.0 & 0.00055; 6.0 & 0.00603; 7.0 & 0.00378; 8.0 & 0.00338; 8.0\\
    RLLS-soft & VS & 0.00171; 9.0 & 0.00086; 8.0 & 0.00056; 10.0 & 0.00626; 8.0 & 0.00405; 11.0 & 0.00393; 13.0\\
  \end{tabular}}
  \caption{\textbf{MNIST: Comparison of all calibration and domain adaptation methods, using MSE (Sec. ~\ref{sec:metrics_for_labelshift}) as the metric (``tweak-one'' shift).} Value before the semicolon is the median of the metric over all trials. Value after the semicolon is the median rank of the domain adaptation + calibration method combination relative to the other method combinations in the column. Bold values in a column are not significantly different from the best-performing method in the column as measured by a paired Wilcoxon test at $p < 0.01$. EM with BCTS or VS tends to achieve the best performance. See \textbf{Sec. ~\ref{sec:experimentalsetup}} for details on the experimental setup.}
  \label{tab:MNIST_all_mseweights_even_tweakoneshift}
\end{table*}

\begin{table*}[!htbp]
\adjustbox{max width=\textwidth}{
  \centering
  \begin{tabular}{ c | c | c c c | c c c | c c c}
    \multirow{2}{*}{\begin{tabular}{c}\textbf{Shift} \\ \textbf{Estimator} \end{tabular}} & \multirow{2}{*}{\begin{tabular}{c}\textbf{Calibration} \\ \textbf{Method} \end{tabular}} & \multicolumn{3}{| c}{$\alpha=0.1$} & \multicolumn{3}{| c}{$\alpha=1.0$} & \multicolumn{3}{| c}{$\alpha=10$}\\ 
    \cline{3-11}
    & & $n$=2000 & $n$=4000 & $n$=8000 & $n$=2000 & $n$=4000 & $n$=8000 & $n$=2000 & $n$=4000 & $n$=8000\\
    \hline
    \hline
    EM & None & \textbf{0.00438; 2.0} & \textbf{0.00294; 2.0} & 0.00227; 2.0 & 0.00339; 4.0 & 0.00256; 4.0 & 0.00207; 4.0 & 0.00217; 4.0 & 0.00133; 4.0 & 0.00093; 4.0\\
    BBSL-hard & None & 0.00491; 2.0 & 0.0036; 2.0 & 0.00244; 2.0 & 0.00276; 3.0 & 0.00167; 2.5 & 0.00125; 3.0 & 0.00208; 3.0 & 0.00106; 3.0 & 0.00073; 3.0\\
    BBSL-soft & None & 0.00459; 2.0 & 0.00324; 2.0 & 0.00213; 2.0 & \textbf{0.00227; 1.0} & \textbf{0.00137; 1.0} & \textbf{0.00085; 1.0} & 0.00174; 1.0 & \textbf{0.00083; 1.0} & \textbf{0.00052; 1.0}\\
    RLLS-hard & None & 0.00472; 2.0 & 0.00347; 2.0 & 0.00238; 2.0 & 0.00271; 2.0 & 0.00167; 2.0 & 0.00125; 2.0 & 0.00208; 2.0 & 0.00106; 2.0 & 0.00073; 2.0\\
    RLLS-soft & None & \textbf{0.0044; 1.0} & \textbf{0.00314; 2.0} & \textbf{0.00211; 1.0} & \textbf{0.00226; 1.0} & \textbf{0.00137; 1.0} & \textbf{0.00085; 1.0} & \textbf{0.00174; 1.0} & \textbf{0.00083; 1.0} & \textbf{0.00052; 0.5}\\
    \hline
    \hline
    EM & TS & \textbf{0.00322; 0.0} & \textbf{0.00195; 0.0} & \textbf{0.00189; 1.0} & 0.00203; 0.0 & 0.00102; 0.0 & 0.00084; 2.0 & 0.00158; 2.0 & 0.00081; 2.0 & 0.00052; 2.0\\
    BBSL-soft & TS & 0.00417; 1.0 & 0.00309; 1.0 & 0.00218; 1.0 & 0.00232; 1.0 & 0.00129; 1.0 & 0.00083; 1.0 & 0.00157; 1.0 & 0.00079; 1.0 & 0.00049; 1.0\\
    RLLS-soft & TS & \textbf{0.00407; 1.0} & \textbf{0.00289; 1.0} & \textbf{0.00206; 1.0} & 0.00232; 1.0 & 0.00129; 1.0 & 0.00083; 1.0 & 0.00157; 1.0 & 0.00079; 1.0 & 0.00049; 1.0\\
    \hline
    \hline
    EM & NBVS & \textbf{0.00144; 0.0} & \textbf{0.00104; 0.0} & \textbf{0.00085; 0.0} & \textbf{0.00167; 0.0} & \textbf{0.00094; 0.0} & \textbf{0.0007; 0.0} & \textbf{0.00165; 1.0} & 0.00081; 0.0 & 0.00051; 0.0\\
    BBSL-soft & NBVS & 0.00428; 1.5 & 0.00306; 1.0 & 0.0022; 1.0 & 0.00234; 1.0 & 0.00129; 1.0 & 0.00082; 1.0 & 0.00171; 1.0 & 0.00085; 1.0 & 0.00052; 1.0\\
    RLLS-soft & NBVS & 0.00409; 1.0 & 0.00284; 1.0 & 0.00211; 1.0 & 0.00232; 1.0 & 0.00129; 1.0 & 0.00082; 1.0 & 0.00171; 1.0 & 0.00085; 1.0 & 0.00052; 1.0\\
    \hline
    \hline
    EM & BCTS & \textbf{0.00085; 0.0} & \textbf{0.00057; 0.0} & \textbf{0.0004; 0.0} & \textbf{0.00165; 0.0} & \textbf{0.00093; 0.0} & 0.0007; 0.0 & \textbf{0.00157; 0.0} & 0.0008; 0.0 & 0.00049; 0.0\\
    BBSL-soft & BCTS & 0.00413; 2.0 & 0.00303; 2.0 & 0.00228; 2.0 & 0.00228; 1.0 & 0.0013; 1.0 & 0.00081; 1.0 & 0.00168; 1.0 & 0.00083; 1.0 & 0.0005; 1.0\\
    RLLS-soft & BCTS & 0.00399; 1.0 & 0.00283; 1.0 & 0.00213; 1.0 & 0.00226; 1.0 & 0.0013; 1.0 & 0.00082; 1.0 & 0.00168; 1.0 & 0.00083; 1.0 & 0.0005; 1.0\\
    \hline
    \hline
    EM & VS & \textbf{0.00087; 0.0} & \textbf{0.00056; 0.0} & \textbf{0.00035; 0.0} & \textbf{0.00172; 0.0} & \textbf{0.00095; 0.0} & \textbf{0.00073; 0.0} & \textbf{0.00156; 0.0} & 0.0008; 0.0 & 0.00054; 0.0\\
    BBSL-soft & VS & 0.0039; 2.0 & 0.00322; 2.0 & 0.00221; 2.0 & 0.0022; 1.0 & 0.00136; 1.0 & 0.00085; 1.0 & 0.00163; 1.0 & 0.00084; 1.0 & 0.00054; 1.0\\
    RLLS-soft & VS & 0.00385; 1.0 & 0.00282; 1.0 & 0.00212; 1.0 & 0.0022; 1.0 & 0.00136; 1.0 & 0.00085; 1.0 & \textbf{0.00163; 1.0} & 0.00084; 1.0 & 0.00054; 1.0\\
  \end{tabular}}
  \caption{\textbf{MNIST: Comparison of EM, BBSL and RLLS (dirichlet shift).} Analogous to \textbf{Table ~\ref{tab:MNIST_emvsbbsevsrlls_mseweights_even_tweakone}}, but with dirichlet shift rather than tweak-one shift.}
  \label{tab:MNIST_emvsbbsevsrlls_mseweights_even_dirichlet}
\end{table*}

\FloatBarrier
\section{CIFAR100 Supplementary Tables}
\FloatBarrier

\begin{table*}[!htbp]
\adjustbox{max width=\textwidth}{
  \centering
  \begin{tabular}{ c | c | c c c | c c c | c c c}
    \multirow{2}{*}{\begin{tabular}{c}\textbf{Shift} \\ \textbf{Estimator} \end{tabular}} & \multirow{2}{*}{\begin{tabular}{c}\textbf{Calibration} \\ \textbf{Method} \end{tabular}} & \multicolumn{3}{| c}{$\alpha=0.1$} & \multicolumn{3}{| c}{$\alpha=1.0$} & \multicolumn{3}{| c}{$\alpha=10.0$}\\ 
    \cline{3-11}
    & & $n$=7000 & $n$=8500 & $n$=10000 & $n$=7000 & $n$=8500 & $n$=10000 & $n$=7000 & $n$=8500 & $n$=10000\\
    \hline
    \hline
    EM & None & 1.80113; 16.0 & 1.67187; 16.0 & 1.7157; 16.0 & 0.55795; 16.0 & 0.54112; 16.0 & 0.55955; 16.0 & 0.3896; 16.0 & 0.37585; 16.0 & 0.36337; 16.0\\
    EM & TS & 0.41127; 9.0 & 0.34963; 7.0 & 0.30451; 6.5 & 0.23803; 13.0 & 0.21715; 13.0 & 0.19781; 13.0 & 0.16408; 12.0 & 0.14477; 13.0 & 0.13852; 13.0\\
    EM & NBVS & 0.1637; 2.0 & 0.15904; 2.0 & 0.14348; 2.0 & 0.1042; 2.0 & 0.10523; 2.0 & 0.10777; 6.0 & 0.10122; 4.5 & 0.09874; 7.0 & 0.09729; 10.0\\
    EM & BCTS & 0.1101; 2.0 & 0.10824; 2.0 & 0.11636; 2.0 & 0.08838; 2.0 & 0.09102; 2.0 & 0.0876; 3.0 & \textbf{0.09207; 2.0} & 0.08707; 5.0 & 0.08781; 8.0\\
    EM & VS & \textbf{0.09485; 1.0} & \textbf{0.08792; 1.0} & \textbf{0.0862; 1.0} & \textbf{0.07684; 1.0} & \textbf{0.07922; 1.0} & \textbf{0.07773; 2.0} & \textbf{0.09387; 2.0} & 0.08796; 5.5 & 0.08981; 9.0\\
    BBSL-hard & None & 0.83095; 15.0 & 0.68099; 15.0 & 0.58656; 15.0 & 0.33542; 15.0 & 0.27998; 15.0 & 0.24659; 15.0 & 0.25694; 15.0 & 0.21736; 15.0 & 0.20637; 15.0\\
    BBSL-soft & None & 0.5279; 13.0 & 0.47521; 12.0 & 0.44263; 13.0 & 0.23637; 13.0 & 0.21324; 13.0 & 0.18605; 13.0 & 0.18842; 13.0 & 0.16108; 13.0 & 0.14018; 13.0\\
    BBSL-soft & TS & 0.40279; 9.0 & 0.34713; 9.0 & 0.33012; 9.0 & 0.18774; 9.5 & 0.15665; 9.0 & 0.12639; 9.0 & 0.13409; 8.5 & 0.10951; 7.0 & 0.09703; 7.0\\
    BBSL-soft & NBVS & 0.40856; 10.0 & 0.33122; 9.0 & 0.29594; 9.0 & 0.17545; 9.0 & 0.1409; 9.0 & 0.11336; 8.0 & 0.13601; 9.0 & 0.113; 8.0 & 0.09731; 7.0\\
    BBSL-soft & BCTS & 0.40887; 9.0 & 0.32629; 9.0 & 0.30435; 9.0 & 0.17923; 9.0 & 0.14756; 9.0 & 0.11536; 9.0 & 0.13698; 9.0 & 0.11267; 8.0 & 0.09748; 7.0\\
    BBSL-soft & VS & 0.43085; 9.0 & 0.3216; 8.0 & 0.30221; 9.0 & 0.16715; 8.0 & 0.13847; 8.0 & 0.11169; 7.0 & 0.13075; 8.0 & 0.10971; 7.0 & 0.09433; 6.0\\
    RLLS-hard & None & 0.4577; 12.0 & 0.38413; 12.0 & 0.37675; 12.0 & 0.20285; 12.0 & 0.16017; 12.0 & 0.14228; 12.0 & 0.16412; 12.0 & 0.13842; 13.0 & 0.12127; 12.0\\
    RLLS-soft & None & 0.33882; 8.0 & 0.2728; 8.0 & 0.2878; 8.0 & 0.13859; 8.0 & 0.12697; 7.0 & 0.10567; 8.0 & 0.12528; 8.0 & 0.11011; 8.5 & 0.10056; 8.0\\
    RLLS-soft & TS & 0.27236; 6.0 & 0.22426; 6.0 & 0.22697; 5.5 & 0.12255; 5.0 & 0.10639; 5.0 & 0.08221; 4.0 & 0.11395; 5.0 & 0.08803; 5.0 & 0.07868; 4.0\\
    RLLS-soft & NBVS & 0.2797; 7.0 & 0.21465; 5.0 & 0.23037; 5.0 & 0.11469; 4.0 & 0.09903; 4.0 & 0.08341; 4.0 & 0.11303; 5.0 & 0.08868; 4.0 & 0.07531; 4.0\\
    RLLS-soft & BCTS & 0.26594; 6.0 & 0.2305; 6.0 & 0.23604; 6.0 & 0.11498; 5.0 & 0.09755; 4.0 & 0.08309; 4.0 & 0.1085; 4.0 & 0.08672; 3.0 & \textbf{0.07252; 3.0}\\
    RLLS-soft & VS & 0.28941; 6.0 & 0.21363; 6.0 & 0.22405; 6.0 & 0.11627; 5.0 & 0.09894; 4.0 & 0.08267; 3.0 & 0.11426; 4.0 & 0.08623; 4.0 & 0.07414; 3.0\\
  \end{tabular}}
  \caption{\textbf{CIFAR100: Comparison of all calibration and domain adaptation methods, using MSE (Sec. ~\ref{sec:metrics_for_labelshift}) as the metric (dirichlet shift).} Value before the semicolon is the median of the metric over all trials. Value after the semicolon is the median rank of the domain adaptation + calibration method combination relative to the other method combinations in the column. Bold values in a column are not significantly different from the best-performing method in the column as measured by a paired Wilcoxon test at $p < 0.01$. EM with VS tends to achieve the best performance, particularly for larger amounts of shift (corresponding to smaller $\alpha$). See \textbf{Sec. ~\ref{sec:experimentalsetup}} for details on the experimental setup.}
  \label{tab:cifar100_all_mseweights_even_dirichletshift}
\end{table*}

\pagebreak
\FloatBarrier
\section{Diabetic Retinopathy Supplementary Tables}
\FloatBarrier

\begin{table*}[!htbp]
\adjustbox{max width=\textwidth}{
  \centering
  \begin{tabular}{ c | c | c c c | c c c}
    \multirow{2}{*}{\begin{tabular}{c}\textbf{Shift} \\ \textbf{Estimator} \end{tabular}} & \multirow{2}{*}{\begin{tabular}{c}\textbf{Calibration} \\ \textbf{Method} \end{tabular}} & \multicolumn{3}{| c}{$\rho=0.5$} & \multicolumn{3}{| c}{$\rho=0.9$}\\ 
    \cline{3-8}
    & & $n$=500 & $n$=1000 & $n$=1500 & $n$=500 & $n$=1000 & $n$=1500\\
    \hline
    \hline
    EM & None & \textbf{0.35073; 3.5} & 0.30498; 4.0 & 0.19709; 3.0 & 0.0899; 10.0 & 0.0596; 12.0 & 0.05666; 13.0\\
    EM & TS & \textbf{0.47048; 3.0} & \textbf{0.29683; 3.0} & 0.20944; 2.0 & 0.08077; 11.0 & 0.05537; 12.0 & 0.05268; 13.0\\
    EM & NBVS & \textbf{0.51515; 5.0} & 0.33256; 4.0 & 0.24066; 4.0 & 0.09137; 12.0 & 0.08415; 14.0 & 0.08002; 15.0\\
    EM & BCTS & \textbf{0.40245; 3.5} & \textbf{0.2475; 3.0} & \textbf{0.19461; 3.0} & \textbf{0.02267; 2.5} & \textbf{0.01354; 2.5} & \textbf{0.01043; 3.0}\\
    EM & VS & 0.53081; 5.0 & \textbf{0.26442; 4.0} & 0.21641; 4.0 & \textbf{0.02388; 3.0} & \textbf{0.01307; 2.0} & \textbf{0.00953; 2.0}\\
    BBSL-hard & None & 1.83879; 15.0 & 1.4584; 15.0 & 0.81962; 15.0 & 0.31856; 16.0 & 0.10582; 15.0 & 0.0491; 13.5\\
    BBSL-soft & None & 1.4041; 11.0 & 0.71345; 10.5 & 0.43408; 10.0 & 0.10441; 12.0 & 0.04494; 11.0 & 0.02562; 10.0\\
    BBSL-soft & TS & 1.26587; 10.0 & 0.57077; 9.0 & 0.38372; 9.0 & 0.09552; 11.0 & 0.03859; 10.0 & 0.02165; 9.0\\
    BBSL-soft & NBVS & 1.62832; 12.0 & 0.70875; 10.5 & 0.45962; 11.0 & 0.09014; 10.0 & 0.03608; 10.0 & 0.01769; 8.0\\
    BBSL-soft & BCTS & 1.40016; 12.0 & 0.52628; 9.0 & 0.46372; 10.0 & 0.05524; 8.0 & 0.02529; 6.0 & 0.01419; 6.0\\
    BBSL-soft & VS & 1.36527; 11.0 & 0.50393; 10.0 & 0.48291; 10.0 & 0.04678; 7.0 & 0.02485; 6.0 & 0.01543; 6.0\\
    RLLS-hard & None & 1.07889; 10.0 & 1.08159; 14.0 & 0.74669; 13.5 & 0.05118; 7.0 & 0.02865; 7.0 & 0.02458; 8.0\\
    RLLS-soft & None & 0.91948; 8.0 & 0.63483; 9.0 & 0.41057; 7.0 & 0.03602; 6.0 & 0.0225; 6.0 & 0.02351; 7.0\\
    RLLS-soft & TS & 0.80812; 7.5 & 0.56293; 8.0 & 0.37624; 8.0 & 0.0352; 6.0 & 0.02457; 7.0 & 0.02054; 7.0\\
    RLLS-soft & NBVS & 0.77171; 7.0 & 0.56492; 7.0 & 0.40272; 9.0 & 0.04301; 7.0 & 0.03258; 8.0 & 0.02541; 8.0\\
    RLLS-soft & BCTS & 0.87398; 8.0 & 0.50668; 7.5 & 0.42255; 9.0 & 0.0384; 6.0 & 0.02826; 6.0 & 0.02245; 6.0\\
    RLLS-soft & VS & 0.82992; 7.0 & 0.47802; 7.0 & 0.42028; 8.5 & 0.04096; 6.0 & 0.02948; 7.0 & 0.022; 6.0\\
  \end{tabular}}
  \caption{\textbf{Kaggle Diabetic Retinopathy Detection: Comparison of all calibration and domain adaptation methods, using MSE (Sec. ~\ref{sec:metrics_for_labelshift}) as the metric.} $\rho$ represents the porportion of healthy examples in the sfhited domain; the source distribution has $\rho=0.73$. Value before the semicolon is the median of the metric over all trials. Value after the semicolon is the median rank of the domain adaptation + calibration method combination relative to the other method combinations in the column. Bold values in a column are not significantly different from the best-performing method in the column as measured by a paired Wilcoxon test at $p < 0.01$. See \textbf{Sec. ~\ref{sec:experimentalsetup}} for details on the experimental setup.}
  \label{tab:kaggledr_all_mseweights_even}
\end{table*}

\FloatBarrier
\section{NLL Corresponds Better To Benefits In Label Shift Adaptation}
\label{sec:nll_corresponds_benefits_labelshift}
\FloatBarrier

To investigate whether NLL or ECE corresponded better to the benefits offered by a calibration method in the context of label shift adaptation, we adopted the following strategy: in a given experimental run, we identified the calibration method that provided the best NLL (or ECE) on the unshifted test set. We then looked at the performance of label shift adaptation using this calibration method. Note that the calibration method selected can differ from one run to the next. Across datasets, we observed that, by and large, selecting a calibration method according to the NLL produced better performance after domain adaptation as compared to selecting a calibration method according to ECE. Results are show in the tables below.

\begin{table*}[!htbp]
\adjustbox{max width=\textwidth}{
  \centering
  \begin{tabular}{ c | c | c c c | c c c | c c c}
    \multirow{2}{*}{\begin{tabular}{c}\textbf{Shift} \\ \textbf{Estimator} \end{tabular}} & \multirow{2}{*}{\begin{tabular}{c}\textbf{Calibration} \\ \textbf{Method} \end{tabular}} & \multicolumn{3}{| c}{$\alpha=0.1$} & \multicolumn{3}{| c}{$\alpha=1.0$} & \multicolumn{3}{| c}{$\alpha=10$}\\ 
    \cline{3-11}
    & & $n$=2000 & $n$=4000 & $n$=8000 & $n$=2000 & $n$=4000 & $n$=8000 & $n$=2000 & $n$=4000 & $n$=8000\\
    \hline
    \hline
    EM & Best NLL & 5.9; 0.0 & 5.775; 0.0 & \textbf{5.75; 0.0} & 2.15; 0.0 & \textbf{2.075; 0.0} & \textbf{2.081; 0.0} & 0.725; 0.0 & \textbf{0.8; 0.0} & \textbf{0.838; 0.0}\\
    EM & Best ECE & 5.675; 1.0 & 5.85; 1.0 & 5.744; 1.0 & 2.15; 1.0 & 1.713; 1.0 & 1.775; 1.0 & 0.75; 1.0 & 0.188; 1.0 & 0.244; 1.0\\
  \end{tabular}}
  \caption{\textbf{CIFAR10: NLL vs ECE, $\Delta$\%Accuracy, dirichlet shift.} Entry in ``calibration method'' column indicates how the calibration method for any given run was selected: either according to whether it produced the best NLL or whether it produced the best ECE, where NLL and ECE were calculated on the unshifted test set. Value before the semicolon is the median change in \%accuracy relative to unadapted predictions. Value after the semicolon is the median rank of the given metric relative to the other metric in the pair. A bold value is significantly better than the non-bold value in the pair using a paired Wilcoxon test at $p \le 0.01$. See \textbf{Sec. ~\ref{sec:experimentalsetup}} for details on the experimental setup.}
  \label{tab:cifar10_nllvsece_deltaacc_dirichletshift}
\end{table*}

\begin{table*}[!htbp]
\adjustbox{max width=\textwidth}{
  \centering
  \begin{tabular}{ c | c | c c c | c c c}
    \multirow{2}{*}{\begin{tabular}{c}\textbf{Shift} \\ \textbf{Estimator} \end{tabular}} & \multirow{2}{*}{\begin{tabular}{c}\textbf{Calibration} \\ \textbf{Method} \end{tabular}} & \multicolumn{3}{| c}{$\rho=0.01$} & \multicolumn{3}{| c}{$\rho=0.9$}\\ 
    \cline{3-8}
    & & $n$=2000 & $n$=4000 & $n$=8000 & $n$=2000 & $n$=4000 & $n$=8000\\
    \hline
    \hline
    EM & Best NLL & \textbf{1.25; 0.0} & \textbf{1.262; 0.0} & \textbf{1.225; 0.0} & 17.3; 0.0 & \textbf{17.025; 0.0} & \textbf{17.362; 0.0}\\
    EM & Best ECE & 1.05; 1.0 & 1.088; 1.0 & 1.137; 1.0 & 17.125; 1.0 & 16.412; 1.0 & 16.45; 1.0\\
  \end{tabular}}
  \caption{\textbf{CIFAR10: NLL vs. ECE, metric: $\Delta$\%accuracy, ``tweak-one'' shift}. Analogous to \textbf{Table ~\ref{tab:cifar10_nllvsece_deltaacc_dirichletshift}}. The ``tweak-one'' shift strategy is explained in \textbf{Sec. ~\ref{sec:experimentalsetup}}.}
  \label{tab:cifar10_nllvsece_deltaacc_tweakoneshift}
\end{table*}

\begin{table*}[!htbp]
\adjustbox{max width=\textwidth}{
  \centering
  \begin{tabular}{ c | c | c c c | c c c | c c c}
    \multirow{2}{*}{\begin{tabular}{c}\textbf{Shift} \\ \textbf{Estimator} \end{tabular}} & \multirow{2}{*}{\begin{tabular}{c}\textbf{Calibration} \\ \textbf{Method} \end{tabular}} & \multicolumn{3}{| c}{$\alpha=0.1$} & \multicolumn{3}{| c}{$\alpha=1.0$} & \multicolumn{3}{| c}{$\alpha=10$}\\ 
    \cline{3-11}
    & & $n$=2000 & $n$=4000 & $n$=8000 & $n$=2000 & $n$=4000 & $n$=8000 & $n$=2000 & $n$=4000 & $n$=8000\\
    \hline
    \hline
    EM & Best NLL & \textbf{0.04654; 0.0} & \textbf{0.03934; 0.0} & \textbf{0.02037; 0.0} & 0.1262; 0.0 & \textbf{0.07123; 0.0} & \textbf{0.04254; 0.0} & 0.1609; 0.5 & \textbf{0.07719; 0.0} & \textbf{0.04652; 0.0}\\
    EM & Best ECE & 0.14529; 1.0 & 0.08345; 1.0 & 0.04786; 1.0 & 0.118; 1.0 & 0.19194; 1.0 & 0.1479; 1.0 & 0.16718; 0.5 & 0.10077; 1.0 & 0.05845; 1.0\\
    \hline
    \hline
    BBSL-soft & Best NLL & 0.28846; 0.0 & 0.13062; 0.0 & 0.10355; 0.0 & 0.17072; 1.0 & 0.10215; 0.5 & 0.06013; 0.0 & 0.17633; 1.0 & \textbf{0.0825; 0.0} & 0.04782; 0.0\\
    BBSL-soft & Best ECE & 0.27553; 1.0 & 0.13608; 1.0 & 0.10031; 1.0 & 0.17342; 0.0 & 0.1021; 0.5 & 0.05966; 1.0 & 0.17173; 0.0 & 0.09483; 1.0 & 0.04756; 1.0\\
    \hline
    \hline
    RLLS-soft & Best NLL & 0.28416; 0.0 & 0.13019; 0.0 & 0.08999; 0.0 & 0.16909; 1.0 & 0.10211; 0.5 & 0.06013; 0.0 & 0.17633; 1.0 & \textbf{0.0825; 0.0} & 0.04782; 0.0\\
    RLLS-soft & Best ECE & 0.27631; 1.0 & 0.13053; 1.0 & 0.09304; 1.0 & 0.17087; 0.0 & 0.1021; 0.5 & 0.05977; 1.0 & 0.17173; 0.0 & 0.09483; 1.0 & 0.04756; 1.0\\
  \end{tabular}}
  \caption{\textbf{CIFAR10: NLL vs. ECE, metric: MSE, dirichlet shift}. Analogous to \textbf{Table ~\ref{tab:cifar10_nllvsece_deltaacc_dirichletshift}}, but using MSE (\textbf{Sec. ~\ref{sec:metrics_for_labelshift}}) as the metric rather than change in \%accuracy.}
  \label{tab:cifar10_nllvsece_mseweights_even_dirichletshift}
\end{table*}

\begin{table*}[!htbp]
\adjustbox{max width=\textwidth}{
  \centering
  \begin{tabular}{ c | c | c c c | c c c}
    \multirow{2}{*}{\begin{tabular}{c}\textbf{Shift} \\ \textbf{Estimator} \end{tabular}} & \multirow{2}{*}{\begin{tabular}{c}\textbf{Calibration} \\ \textbf{Method} \end{tabular}} & \multicolumn{3}{| c}{$\rho=0.01$} & \multicolumn{3}{| c}{$\rho=0.9$}\\ 
    \cline{3-8}
    & & $n$=2000 & $n$=4000 & $n$=8000 & $n$=2000 & $n$=4000 & $n$=8000\\
    \hline
    \hline
    EM & Best NLL & \textbf{0.13671; 0.0} & \textbf{0.05747; 0.0} & \textbf{0.03794; 0.0} & 0.12478; 0.0 & \textbf{0.08311; 0.0} & \textbf{0.07207; 0.0}\\
    EM & Best ECE & 0.14919; 1.0 & 0.06362; 1.0 & 0.04005; 1.0 & 0.16472; 1.0 & 4.89671; 1.0 & 4.23677; 1.0\\
    \hline
    \hline
    BBSL-soft & Best NLL & 0.15212; 0.0 & 0.07357; 1.0 & 0.04304; 1.0 & 1.18879; 1.0 & \textbf{0.60651; 0.0} & \textbf{0.35089; 0.0}\\
    BBSL-soft & Best ECE & 0.14834; 1.0 & \textbf{0.07226; 0.0} & \textbf{0.04296; 0.0} & 1.21037; 0.0 & 0.85178; 1.0 & 0.5647; 1.0\\
    \hline
    \hline
    RLLS-soft & Best NLL & 0.15212; 0.0 & 0.07357; 1.0 & 0.04304; 1.0 & 1.17799; 1.0 & \textbf{0.60084; 0.0} & \textbf{0.337; 0.0}\\
    RLLS-soft & Best ECE & 0.14834; 1.0 & \textbf{0.07226; 0.0} & \textbf{0.04296; 0.0} & 1.21754; 0.0 & 0.83205; 1.0 & 0.5647; 1.0\\
  \end{tabular}}
  \caption{\textbf{CIFAR10: NLL vs ECE, metric: MSE, ``tweak-one'' shift.} Analogous to \textbf{Table ~\ref{tab:cifar10_nllvsece_deltaacc_dirichletshift}}.}
  \label{tab:cifar10_nllvsece_mseweights_even_tweakoneshift}
\end{table*}

\begin{table*}[!htbp]
\adjustbox{max width=\textwidth}{
  \centering
  \begin{tabular}{ c | c | c c c | c c c | c c c}
    \multirow{2}{*}{\begin{tabular}{c}\textbf{Shift} \\ \textbf{Estimator} \end{tabular}} & \multirow{2}{*}{\begin{tabular}{c}\textbf{Calibration} \\ \textbf{Method} \end{tabular}} & \multicolumn{3}{| c}{$\alpha=0.1$} & \multicolumn{3}{| c}{$\alpha=1.0$} & \multicolumn{3}{| c}{$\alpha=10.0$}\\ 
    \cline{3-11}
    & & $n$=7000 & $n$=8500 & $n$=10000 & $n$=7000 & $n$=8500 & $n$=10000 & $n$=7000 & $n$=8500 & $n$=10000\\
    \hline
    \hline
    EM & Best NLL & \textbf{24.829; 0.0} & \textbf{24.518; 0.0} & 24.51; 0.0 & \textbf{21.3; 0.0} & \textbf{21.629; 0.0} & \textbf{21.93; 0.0} & \textbf{21.229; 0.0} & \textbf{21.306; 0.0} & \textbf{21.21; 0.0}\\
    EM & Best ECE & 24.079; 1.0 & 24.053; 1.0 & 24.575; 1.0 & 20.986; 1.0 & 21.382; 1.0 & 21.53; 1.0 & 21.05; 1.0 & 20.971; 1.0 & 21.06; 1.0\\
  \end{tabular}}
  \caption{\textbf{CIFAR100: NLL vs ECE, metric: $\Delta$\%Accuracy, dirichlet shift.} Analogous to \textbf{Table ~\ref{tab:cifar10_nllvsece_deltaacc_dirichletshift}}}
  \label{tab:cifar100_nllvsece_deltaacc_dirichletshift}
\end{table*}

\begin{table*}[!htbp]
\adjustbox{max width=\textwidth}{
  \centering
  \begin{tabular}{ c | c | c c c | c c c | c c c}
    \multirow{2}{*}{\begin{tabular}{c}\textbf{Shift} \\ \textbf{Estimator} \end{tabular}} & \multirow{2}{*}{\begin{tabular}{c}\textbf{Calibration} \\ \textbf{Method} \end{tabular}} & \multicolumn{3}{| c}{$\alpha=0.1$} & \multicolumn{3}{| c}{$\alpha=1.0$} & \multicolumn{3}{| c}{$\alpha=10.0$}\\ 
    \cline{3-11}
    & & $n$=7000 & $n$=8500 & $n$=10000 & $n$=7000 & $n$=8500 & $n$=10000 & $n$=7000 & $n$=8500 & $n$=10000\\
    \hline
    \hline
    EM & Best NLL & \textbf{0.09485; 0.0} & \textbf{0.08792; 0.0} & \textbf{0.0862; 0.0} & \textbf{0.07684; 0.0} & \textbf{0.07922; 0.0} & \textbf{0.07773; 0.0} & \textbf{0.09307; 0.0} & \textbf{0.0868; 0.0} & \textbf{0.08981; 0.0}\\
    EM & Best ECE & 0.1637; 1.0 & 0.15904; 1.0 & 0.14348; 1.0 & 0.1042; 1.0 & 0.10523; 1.0 & 0.10777; 1.0 & 0.10042; 1.0 & 0.0966; 1.0 & 0.09683; 1.0\\
    \hline
    \hline
    BBSL-soft & Best NLL & 0.43085; 0.0 & \textbf{0.3216; 0.0} & 0.30221; 0.0 & \textbf{0.16715; 0.0} & \textbf{0.13847; 0.0} & \textbf{0.11169; 0.0} & \textbf{0.13481; 0.0} & \textbf{0.11224; 0.0} & \textbf{0.09433; 0.0}\\
    BBSL-soft & Best ECE & 0.40856; 1.0 & 0.33122; 1.0 & 0.29594; 1.0 & 0.17545; 1.0 & 0.1409; 1.0 & 0.11336; 1.0 & 0.13601; 1.0 & 0.113; 1.0 & 0.09731; 1.0\\
    \hline
    \hline
    RLLS-soft & Best NLL & 0.28941; 0.5 & 0.21363; 1.0 & 0.22405; 1.0 & 0.11627; 0.0 & 0.09894; 1.0 & 0.08267; 0.0 & 0.11264; 0.0 & 0.08623; 0.0 & \textbf{0.07414; 0.0}\\
    RLLS-soft & Best ECE & 0.2797; 0.5 & 0.21465; 0.0 & 0.23037; 0.0 & 0.11469; 1.0 & 0.09903; 0.0 & 0.08341; 1.0 & 0.11303; 1.0 & 0.08868; 1.0 & 0.07531; 1.0\\
  \end{tabular}}
  \caption{\textbf{CIFAR100: NLL vs ECE, metric: MSE, dirichlet shift.} Analogous to \textbf{Table ~\ref{tab:cifar10_nllvsece_deltaacc_dirichletshift}}}
  \label{tab:cifar100_nllvsece_mseweights_even_dirichletshift}
\end{table*}

\begin{table*}[!htbp]
\adjustbox{max width=\textwidth}{
  \centering
  \begin{tabular}{ c | c | c c c | c c c}
    \multirow{2}{*}{\begin{tabular}{c}\textbf{Shift} \\ \textbf{Estimator} \end{tabular}} & \multirow{2}{*}{\begin{tabular}{c}\textbf{Calibration} \\ \textbf{Method} \end{tabular}} & \multicolumn{3}{| c}{$\rho=0.5$} & \multicolumn{3}{| c}{$\rho=0.9$}\\ 
    \cline{3-8}
    & & $n$=500 & $n$=1000 & $n$=1500 & $n$=500 & $n$=1000 & $n$=1500\\
    \hline
    \hline
    EM & Best NLL & \textbf{3.8; 0.0} & \textbf{4.5; 0.0} & \textbf{4.6; 0.0} & \textbf{3.6; 0.0} & 3.6; 0.0 & \textbf{3.733; 0.0}\\
    EM & Best ECE & 3.6; 1.0 & 4.4; 1.0 & 4.3; 1.0 & 2.2; 1.0 & 3.6; 1.0 & 2.467; 1.0\\
  \end{tabular}}
  \caption{\textbf{KaggleDR: NLL vs ECE, metric: $\Delta$\%Accuracy.} Shift strategy modifies the proportion of healthy examples. Analogous to \textbf{Table ~\ref{tab:cifar10_nllvsece_deltaacc_dirichletshift}}}
  \label{tab:kaggledr_nllvsece_deltaperf}
\end{table*}

\begin{table*}[!htbp]
\adjustbox{max width=\textwidth}{
  \centering
  \begin{tabular}{ c | c | c c c | c c c}
    \multirow{2}{*}{\begin{tabular}{c}\textbf{Shift} \\ \textbf{Estimator} \end{tabular}} & \multirow{2}{*}{\begin{tabular}{c}\textbf{Calibration} \\ \textbf{Method} \end{tabular}} & \multicolumn{3}{| c}{$\rho=0.5$} & \multicolumn{3}{| c}{$\rho=0.9$}\\ 
    \cline{3-8}
    & & $n$=500 & $n$=1000 & $n$=1500 & $n$=500 & $n$=1000 & $n$=1500\\
    \hline
    \hline
    EM & Best NLL & 0.437; 0.0 & \textbf{0.242; 0.0} & 0.21; 0.0 & \textbf{0.023; 0.0} & 0.013; 0.0 & \textbf{0.01; 0.0}\\
    EM & Best ECE & 0.485; 1.0 & 0.329; 1.0 & 0.225; 1.0 & 0.091; 1.0 & 0.013; 1.0 & 0.08; 1.0\\
    \hline
    \hline
    BBSL-soft & Best NLL & 1.4; 0.0 & \textbf{0.554; 0.0} & 0.486; 0.0 & \textbf{0.055; 0.0} & 0.025; 0.0 & \textbf{0.015; 0.0}\\
    BBSL-soft & Best ECE & 1.498; 1.0 & 0.682; 1.0 & 0.416; 1.0 & 0.09; 1.0 & 0.025; 1.0 & 0.018; 1.0\\
    \hline
    \hline
    RLLS-soft & Best NLL & 0.907; 0.0 & 0.5; 0.0 & 0.404; 0.0 & 0.038; 0.0 & 0.029; 0.0 & \textbf{0.022; 0.0}\\
    RLLS-soft & Best ECE & \textbf{0.835; 1.0} & 0.551; 1.0 & 0.403; 1.0 & 0.043; 1.0 & 0.029; 1.0 & 0.025; 1.0\\
  \end{tabular}}
  \caption{\textbf{KaggleDR: NLL vs ECE, metric: MSE.} Shift strategy modifies the proportion of healthy examples. Analogous to \textbf{Table ~\ref{tab:cifar10_nllvsece_deltaacc_dirichletshift}}}
  \label{tab:kaggledr_nllvsece_mseweights_even}
\end{table*}

\pagebreak
\section{Proof that EM Converges to a Stationary Point of the Likelihood Function}
\label{sec:proof_convergence_stationary}

We slightly modify the derivation of Theorem 2 of Wu (1983), which states that (given certain conditions) continuity of the conditional expectation $Q(q(\boldsymbol{\omega}) ; q^{(s)}(\boldsymbol{\omega}) )$ in $q(\boldsymbol{\omega})$ and $q^{(s)}(\boldsymbol{\omega})$ implies that the sequence of parameter estimates produced by EM converges to a stationary point of the likelihood. The reason we do not directly invoke Theorem 2 of Wu (1983) is that its proof assumes that the conditional expectation $Q(q(\boldsymbol{\omega}) ; q^{(s)}(\boldsymbol{\omega}) )$ is defined at all points in the domain of the log-likelihood. In our application, $Q(q(\boldsymbol{\omega}) ; q^{(s)}(\boldsymbol{\omega}) )$ can be undefined when $q(\boldsymbol{\omega})$ is at the boundary of the domain - however, in spite of this, we observe that it is still possible to prove that EM converges to a stationary point of the log-likelihood. 

The proof proceeds as follows: in \textbf{Sec. \ref{sec:review_likelihood}}, we review the definition of the likelihood and recap key properties of $Q(q(\boldsymbol{\omega}) ; q^{(s)}(\boldsymbol{\omega}) )$. In \textbf{Sec. \ref{sec:verifying_conditions}}, we verify that certain conditions assumed in Wu (1983) hold for the case of domain adaptation to label shift. In \textbf{Sec. \ref{sec:proving_smoothness_of_Q}}, we identify the domain of $Q(q(\boldsymbol{\omega}) ; q^{(s)}(\boldsymbol{\omega}) )$ and prove differentiability over this domain. Finally, in \textbf{Sec. \ref{sec:invoke_theorem_1}}, we will invoke Theorem 1 of Wu (1983) to show that EM converges to a stationary point of the likelihood. Together with the concavity of the likelihood, this is sufficient to guarantee convergence to a global maximum.

\subsection{Review of Likelihood and Key Properties of $Q(q(\boldsymbol{\omega}) ; q^{(s)}(\boldsymbol{\omega}) )$}
\label{sec:review_likelihood}

We will begin by recapping certain definitions and expressions that will be useful later in the proof. Let $\omega_i$ denote membership in class $i$, and let $q(\omega_i)$ and $p(\omega_i)$ denote the target \& source domain priors. Using the form of the log-likelihood derived in \textbf{Eqn. \ref{eqn:rewrittenemobjective}} in the main text, we have:

\begin{align}
    L(q(\boldsymbol{\omega})) &= \sum_k \log \left( \sum_i q(\boldsymbol{x}_k | \omega_i)  q(\omega_i) \right) \nonumber \\
    &= \sum_k \log \left( \sum_i p(\boldsymbol{x}_k | \omega_i)  q(\omega_i) \right) \text{ (By the label shift assumption)} \nonumber
\end{align}

Let $q^{(s)}(\omega_i | \boldsymbol{x_k})$ denote the conditional probability that $\boldsymbol{x}_k$ originated from class $i$ given the estimate $q^{(s)}(\boldsymbol{\omega})$ of the priors at EM iteration $s$. The expression for $q^{(s)}(\omega_i | \boldsymbol{x_k})$ in terms of $q^{(s)}(\omega_i)$ can be found in equation A.4 of Saerens 2002:
\begin{align}
    q^{(s)}(\omega_i|\boldsymbol{x}_k) = \frac{\frac{q^{(s)}(\omega_i)}{p(\omega_i)}p(\omega_i|\boldsymbol{x}_k)}{\sum_{j=1}^{m}\frac{q^{(s)}(\omega_j)}{p(\omega_j)}p(\omega_j|\boldsymbol{x}_k)}= \frac{q^{(s)}(\omega_i)p(\omega_i|\boldsymbol{x}_k)}{p(\omega_i)\sum_{j=1}^{m}\frac{q^{(s)}(\omega_j)}{p(\omega_j)}p(\omega_j|\boldsymbol{x}_k)} \label{eqn:qs_omega_given_x}
\end{align}

Following the standard derivation of EM, we can rewrite the log-likelihood as:
\begin{align}
    L(q(\boldsymbol{\omega}))
    &= \sum_k\log \left( \sum_i \frac{q^{(s)}(\omega_i | \boldsymbol{x_k})}{q^{(s)}(\omega_i | \boldsymbol{x_k})} p(\boldsymbol{x}_k | \omega_i) q(\omega_i) \right) \nonumber\\
    &= \sum_k\log \left( \mathbbm{E}_{i \sim q^{(s)}(\omega_i | \boldsymbol{x_k})} \left[ \frac{1}{q^{(s)}(\omega_i | \boldsymbol{x}_k)} p(\boldsymbol{x}_k | \omega_i) q(\omega_i) \right] \right) \label{eqn:prejensens}\\
    &\ge \sum_k \mathbbm{E}_{i \sim q^{(s)}(\omega_i | \boldsymbol{x_k})} \left[ \log \left( \frac{1}{q^{(s)}(\omega_i | \boldsymbol{x_k})} p(\boldsymbol{x}_k | \omega_i) q(\omega_i) \right) \right] \text{ (By Jensen's inequality)} \nonumber\\
    &= \sum_k \sum_i q^{(s)}(\omega_i | \boldsymbol{x_k}) \log \left( \frac{1}{q^{(s)}(\omega_i | \boldsymbol{x_k})} p(\boldsymbol{x}_k | \omega_i) q(\omega_i) \right) \nonumber\\
    &= \sum_k \sum_i q^{(s)}(\omega_i | \boldsymbol{x_k}) \log \left( p(\boldsymbol{x}_k | \omega_i) q(\omega_i) \right) - \sum_k \sum_i q^{(s)}(\omega_i | \boldsymbol{x_k}) \log \left( q^{(s)}(\omega_i | \boldsymbol{x_k}) \right) \nonumber\\
\nonumber
\end{align}

Let us define:
\begin{equation}
Q(q(\boldsymbol{\omega}); q^{(s)}(\boldsymbol{\omega}) ) := \sum_k \sum_i q^{(s)}(\omega_i | \boldsymbol{x_k}) \log \left( p(\boldsymbol{x}_k | \omega_i) q(\omega_i) \right) \label{eqn:definition_of_Q}
\end{equation}
This agrees with the definition of $Q(q(\boldsymbol{\omega}); q^{(s)}(\boldsymbol{\omega}) )$ as the conditional expectation of the complete data likelihood given $q^{(s)}(\boldsymbol{\omega})$; for a derivation, see equation A.5 in Saerens (2002).

Given our definition of $Q(q(\boldsymbol{\omega}); q^{(s)}(\boldsymbol{\omega}) )$, we can write:
\begin{align}
&L(q(\boldsymbol{\omega})) \ge Q(q(\boldsymbol{\omega}); q^{(s)}(\boldsymbol{\omega}) ) - \sum_k \sum_i q^{(s)}(\omega_i | \boldsymbol{x_k}) \log \left( q^{(s)}(\omega_i | \boldsymbol{x_k}) \right) \nonumber \\
&\sum_k \sum_i q^{(s)}(\omega_i | \boldsymbol{x_k}) \log \left( q^{(s)}(\omega_i | \boldsymbol{x_k}) \right) \ge  Q(q(\boldsymbol{\omega}); q^{(s)}(\boldsymbol{\omega}) ) - L(q(\boldsymbol{\omega}))  \nonumber
\end{align}

Let us define
\begin{equation}
H(q(\boldsymbol{\omega}); q^{(s)}(\boldsymbol{\omega}) ) :=  Q(q(\boldsymbol{\omega}); q^{(s)}(\boldsymbol{\omega}) ) - L(q(\boldsymbol{\omega}))
\end{equation}
We have:
\begin{equation}
\sum_k \sum_i q^{(s)}(\omega_i | \boldsymbol{x_k}) \log \left( q^{(s)}(\omega_i | \boldsymbol{x_k}) \right) \ge H(q(\boldsymbol{\omega}); q^{(s)}(\boldsymbol{\omega}) )
\end{equation}

When $q^{(s)}(\boldsymbol{\omega}) = q(\boldsymbol{\omega})$, the expectation $\mathbbm{E}_{i \sim q^{(s)}(\omega_i | \boldsymbol{x_k})} \left[ \frac{1}{q^{(s)}(\omega_i)} p(\boldsymbol{x}_k | \omega_i) q(\omega_i) \right]$ in (\ref{eqn:prejensens}) becomes  $\mathbbm{E}_{i \sim q^{(s)}(\omega_i | \boldsymbol{x_k})} \left[p(\boldsymbol{x}_k | \omega_i)\right]$, which is the expectation over a constant-valued random variable.  Because Jensen's inequality holds with equality when the expectation is taken over a constant-valued random variable, we have:
\begin{align}
\sum_k \sum_i q^{(s)}(\omega_i | \boldsymbol{x_k}) \log \left( q^{(s)}(\omega_i | \boldsymbol{x_k}) \right) = H(q^{(s)}(\boldsymbol{\omega}); q^{(s)}(\boldsymbol{\omega}) )
\end{align}

Because $ \sum_k \sum_i q^{(s)}(\omega_i | \boldsymbol{x_k}) \log \left( q^{(s)}(\omega_i | \boldsymbol{x_k}) \right)$ does not depend on $q(\boldsymbol{\omega})$, this implies $H(q(\boldsymbol{\omega}); q^{(s)}(\boldsymbol{\omega}) )$ is maximized when $q^{(s)}(\boldsymbol{\omega}) = q(\boldsymbol{\omega})$, i.e.

\begin{align}
q^{(s)}(\boldsymbol{\omega}) = \argmax_{q(\boldsymbol{\omega})} H(q(\boldsymbol{\omega}); q^{(s)}(\boldsymbol{\omega}) ) \label{eqn:qsw_maximises_H}
\end{align}

It follows that if we set $q^{(s+1)}(\boldsymbol{\omega})$ to be $\argmax_{q(\boldsymbol{\omega})} Q(q(\boldsymbol{\omega}); q^{(s)}(\boldsymbol{\omega}) )$, we are guaranteed that $L(q^{(s+1)}) \ge L(q^{(s)})$:
\begin{align}
    L(q^{(s+1)}(\boldsymbol{\omega})) &= Q(q^{(s+1)}(\boldsymbol{\omega}); q^{(s)}(\boldsymbol{\omega})) - H(q^{(s+1)}(\boldsymbol{\omega}); q^{(s)}(\boldsymbol{\omega})) \nonumber\\
    &\ge Q(q^{(s)}(\boldsymbol{\omega}); q^{(s)}(\boldsymbol{\omega})) - H(q^{(s+1)}(\boldsymbol{\omega}); q^{(s)}(\boldsymbol{\omega})) \text{ (From the definition of } q^{(s+1)}(\boldsymbol{\omega})\text{)}\nonumber\\
    &\ge Q(q^{(s)}(\boldsymbol{\omega}); q^{(s)}(\boldsymbol{\omega})) - H(q^{(s)}(\boldsymbol{\omega}); q^{(s)}(\boldsymbol{\omega})) \text{ (From \textbf{Eqn. \ref{eqn:qsw_maximises_H}})} \nonumber\\
    &= L(q^{(s)}(\boldsymbol{\omega})) \label{eqn:em_monotonic_improvement}
\end{align}

The formula for the EM update rule $q^{(s+1)}(\boldsymbol{\omega}) = \argmax_{q(\boldsymbol{\omega})} Q(q(\boldsymbol{\omega}); q^{(s)}(\boldsymbol{\omega}) )$ can be found in Equation A.8 of Saerens (2002) and is reproduced below:

\begin{equation}
    q^{(s+1)}(\omega_i) = \frac{1}{N} \sum_{k=1}^N q^{(s)} (\omega_i | \boldsymbol{x}_k) \label{eqn:em_update_rule}
\end{equation}
where $N$ denotes the number of examples from the target domain and $q^{(s)} (\omega_i | \boldsymbol{x}_k)$ is defined as in (\ref{eqn:qs_omega_given_x}).

\subsection{Verifying Conditions Assumed In Wu (1983)}
\label{sec:verifying_conditions}

Given our assumption that $p(\omega_i) \geq \epsilon $ $\forall i$, we will show that the conditions assumed in Wu (1983) hold for domain adaptation to label shift. Let $\boldsymbol{\Omega}$ denote the domain of the log-likelihood function $L(q(\boldsymbol{\omega})) = C + \sum_k\log \left( \sum_i \frac{p(\omega_i|\boldsymbol{x}_k)}{p(\omega_i)} q(\omega_i) \right)$. The conditions are:
\begin{enumerate}
    \item $\boldsymbol{\Omega}$ is a subset in $d$-dimensional Euclidean space $\mathbbm{R}^d$
    \item $\boldsymbol{\Omega}_{q_o(\boldsymbol{\omega})} = \{ q(\boldsymbol{\omega}) \in \boldsymbol{\Omega} : L(q(\boldsymbol{\omega})) \ge L(q_o(\boldsymbol{\omega})) \}$ is compact for any $L(q_o(\boldsymbol{\omega})) > -\infty$
    \item $L(q(\boldsymbol{\omega}))$ is continuous in $\boldsymbol{\Omega}$ and differentiable in the interior of $\boldsymbol{\Omega}$
    \item The EM starting point $q^{(0)}(\boldsymbol{\omega})$ satisfies $L(q^{(0)}(\boldsymbol{\omega})) > -\infty$
    \item Each $q^{(s)}(\boldsymbol{\omega})$ (denoting the parameters at EM iteration $s$) lies in the interior of $\boldsymbol{\Omega}$
\end{enumerate}

We verify each condition in turn.

\subsubsection{$\boldsymbol{\Omega}$ is a subset in $d$-dimensional Euclidean space $\mathbbm{R}^d$}
\label{sec:domainoflikelihooddefinition}

\textbf{Proof:} let $d$ denote the number of classes. Given our assumption that $p(\omega_i) \ge \epsilon$ $\forall i$, we observe that the log likelihood is defined so long as $\sum_{i=1}^d p(\omega_i | \boldsymbol{x}_k)q(\omega_i) > 0$ $\forall k$. Thus, the domain $\boldsymbol{\Omega}$ is $\{q(\boldsymbol{\omega}) : q(\omega_i) \ge 0$ $\forall i, \sum_{i=1}^d q(\omega_i) = 1, \sum_{i=1}^d p(\omega_i|\boldsymbol{x}_k)q(\omega_i) > 0$ $\forall k\}$, which is a subset of $\mathbbm{R}^d$ (more specifically, a subset of the hyperplane defined by $\sum_{i=1}^d q(\omega_i) = 1$).\QED

\subsubsection{$\boldsymbol{\Omega}_{q_o(\boldsymbol{\omega})} = \{ q(\boldsymbol{\omega}) \in \boldsymbol{\Omega} : L(q(\boldsymbol{\omega})) \ge L(q_o(\boldsymbol{\omega})) \}$ is compact for any $L(q_o(\boldsymbol{\omega})) > -\infty$}

\textbf{Proof:} for a subset of Euclidean space $\mathbbm{R}^d$ to be compact, it must be both bounded and closed. The parameter space is bounded because $0 \le q(\omega_i) \le 1$ $\forall i$. To show closedness, we will use proof by contradiction. Assume that there exists some $q(\boldsymbol{\omega})$ s.t. $\boldsymbol{\Omega}_{q_o(\boldsymbol{\omega})}$ is open and $L(q_o(\boldsymbol{\omega})) > -\infty$. We begin by noting that $L(q(\boldsymbol{\omega}))$ is finite everywhere in $\boldsymbol{\Omega}$ (by definition of $\boldsymbol{\Omega}$), and is also continuous everywhere in $\boldsymbol{\Omega}$ (proven in \textbf{Sec. \ref{sec:proof_L_continuous_smooth}}). If $\boldsymbol{\Omega}_{q_o(\boldsymbol{\omega})}$ were open, it would imply that $\exists x \notin \boldsymbol{\Omega}_{q_o(\boldsymbol{\omega})}$ and a sequence $\{x'_n\} \in \boldsymbol{\Omega}_{q_o(\boldsymbol{\omega})}$ and $x'_n \rightarrow x$. From continuity of $L(q(\boldsymbol{\omega}))$ over $\boldsymbol{\Omega}$, we know that if $x \in \boldsymbol{\Omega}$, then $L(x'_n) \ge L(q_o(\boldsymbol{\omega}))$ $\forall n$ as $x'_n \rightarrow x$ implies that $L(x) \ge L(q_o(\boldsymbol{\omega}))$. In other words, continuity necessitates that if $x \in \boldsymbol{\Omega}$ and $x'_n \in \boldsymbol{\Omega}_{q_o(\boldsymbol{\omega})}$ $\forall n$, then $x \in \boldsymbol{\Omega}_{q_o(\boldsymbol{\omega})}$. Openness of $\boldsymbol{\Omega}_{q_o(\boldsymbol{\omega})}$ is therefore only possible if $\exists x,\{x'_n\}$ s.t. $x \notin \boldsymbol{\Omega}$ and $x'_n \in \boldsymbol{\Omega}_{q_o(\boldsymbol{\omega})}$ $\forall n$ - in other words, $x$ must exist at an open boundary of $\boldsymbol{\Omega}$. However, the only open boundaries of $\boldsymbol{\Omega}$ are $\{q(\boldsymbol{\omega}): \exists k$ s.t. $\sum_i p(\omega_i | \boldsymbol{x}_k)q(\omega_i)=0 \}$. We observe that $L(q(\boldsymbol{\omega})) \rightarrow -\infty$ as $\sum_i p(\omega_i | \boldsymbol{x}_k)q(\omega_i) \rightarrow 0$ for any $k$. The set $\boldsymbol{\Omega}_{q_o(\boldsymbol{\omega})}$ can therefore only be open if $L(q_o(\boldsymbol{\omega})) = -\infty$, leading to a contradiction. 
\QED

\subsubsection{$L(q(\boldsymbol{\omega}))$ is continuous in $\boldsymbol{\Omega}$ and differentiable in the interior of $\boldsymbol{\Omega}$}
\label{sec:proof_L_continuous_smooth}

\textbf{Proof:} because differentiability everywhere in the domain implies continuity in the domain, it suffices to show that $L(q(\boldsymbol{\omega}))$ is differentiable at all points in the domain. The partial derivative of the log-likelihood $L$ is:
\begin{equation}
    \frac{\partial L(q(\boldsymbol{\omega}))}{\partial q(\omega_i)} = \sum_k \frac{ p(\omega_i|\boldsymbol{x}_k) }{p(\omega_i) \left(\sum_j \frac{p(\omega_j|\boldsymbol{x}_k)}{p(\omega_j)} q(\omega_j) \right) }
\end{equation}
Because we assumed $p(\omega_i) \ge \epsilon$ $\forall i$, we note that the derivative is defined as long as $\sum_j p(\omega_j | \boldsymbol{x}_k) q(\omega_j) > 0$ $\forall k$. As the latter condition is a requirement for being in the domain $\boldsymbol{\Omega}$ (see \textbf{Sec. \ref{sec:domainoflikelihooddefinition}}), we conclude that the log-likelihood $L$ is both continuous and differentiable everywhere in $\boldsymbol{\Omega}$. \QED

\subsubsection{The EM starting point $q^{(0)}(\boldsymbol{\omega})$ satisfies $L(q^{(0)}(\boldsymbol{\omega})) > -\infty$}

\textbf{Proof:} we set $q^{(0)}$ to be equal to the source domain probabilities $p(\omega_i)$. Substituting $q(\omega_i)=p(\omega_i)$ into the expression for the log-likelihood gives:
\begin{align*}
    L(q^{(0)}(\boldsymbol{\omega})) &= C + \sum_k \log\left(\sum_i \frac{p(\omega_i | \boldsymbol{x}_k)}{p(\omega_i)} p(\omega_i)\right)\\
    &= C + \sum_k \log\left(\sum_i p(\omega_i | \boldsymbol{x}_k) \right)\\
    &= C
\end{align*}
As a reminder, $C = \sum_k \log(p(\boldsymbol{x}_k))$ is a constant w.r.t. $q(\boldsymbol{\omega})$ and is therefore ignored when optimizing w.r.t. $q(\boldsymbol{\omega})$. \QED

\subsubsection{Each $q^{(s)}(\boldsymbol{\omega})$ (denoting the parameters at EM iteration $s$) lies in the interior of $\boldsymbol{\Omega}$}
\label{sec:qs_interior_of_omega}

We first state what it means to for $q^{(s)}(\boldsymbol{\omega})$ to lie in the interior of $\boldsymbol{\Omega}$. Because $\boldsymbol{\Omega}$ is a subset of the Euclidean hyperplane defined by $\sum_i q(\omega_i) = 1$, the interior of $\boldsymbol{\Omega}$ consists of $\{q(\boldsymbol{\omega}) : q(\omega_i) > 0$ $\forall i$ and $\sum_i q(\omega_i) = 1\}$. Similarly, the boundary of $\boldsymbol{\Omega}$ consists of $\{q(\boldsymbol{\omega}) : \exists i $ s.t. $q(\omega_i)=0$ and $\sum_i q(\omega_i) = 1\}$. We will use $\partial \boldsymbol{\Omega}$ to denote the boundary of $\boldsymbol{\Omega}$ and $(\boldsymbol{\Omega} \setminus \partial \boldsymbol{\Omega})$ to denote the interior.

We now consider the condition that $q^{(s)}(\boldsymbol{\omega}) \in (\boldsymbol{\Omega} \setminus \partial\boldsymbol{\Omega})$ $\forall s$. We first show that this condition is satisfied when $\sum_k p(\omega_{i} | \boldsymbol{x}_k) > 0$ $\forall i$.\\

\textbf{Lemma:} if $\sum_k p(\omega_{i} | \boldsymbol{x}_k) > 0$ $\forall i$ and $p(\omega_i) \ge \epsilon$ $\forall i$, then $q^{(s)}(\omega_i) > 0$ $\forall i, s$.\\
\textbf{Proof:} Note that $\sum_k p(\omega_{i} | \boldsymbol{x}_k) > 0$ implies that $\exists k$ s.t. $p(\omega_{i} | \boldsymbol{x}_k) > 0$. We use proof by induction. The base case is satisfied because $q^{(0)}(\boldsymbol{\omega})$ is initialized to $p(\boldsymbol{\omega})$, which (by assumption) satisfies $p(\omega_i) \ge \epsilon$ $\forall i$. We now observe that the product $q^{(s)}(\omega_i)p(\omega_{i} | \boldsymbol{x}_k)$ is the numerator of $q^{(s)}(\omega_i | \boldsymbol{x}_k) = \frac{q^{(s)}(\omega_i)p(\omega_{i} | \boldsymbol{x}_k)}{p(\omega_i) \sum_{j=1}^m q^{(s)}(\omega_j)p(\omega_{j} | \boldsymbol{x}_k)}$; thus, $q^{(s)}(\omega_i | \boldsymbol{x}_k) > 0$ if $q^{(s)}(\omega_i) > 0$ and $p(\omega_{i} | \boldsymbol{x}_k) > 0$. It therefore follows that $q^{(s+1)}(\omega_i) = \frac{1}{N} \sum_{k=1}^N q^{(s)}(\omega_i | \boldsymbol{x}_k) > 0$ if $q^{(s)}(\omega_i)$ and $\exists k$ s.t. $p(\omega_{i} | \boldsymbol{x}_k) > 0$. \QED \\

We now consider the case where $\exists i'$ s.t. $\sum_k p(\omega_{i'} | \boldsymbol{x}_k) = 0$. In this case, the condition is not technically satisfied because $q^{(s)}(\omega_{i'} | \boldsymbol{x}_k) = 0$ when $p(\omega_{i'} | \boldsymbol{x}_k) = 0$, and thus $q^{(s+1)}(\omega_{i'}) = \frac{1}{N} \sum_{k=1}^N q^{(s)}(\omega_{i'} | \boldsymbol{x}_k)$ would be $0$ for all $k+1 > 0$. However, the EM updates in this case would be equivalent to performing EM on a reduced problem where class $i'$ is simply excluded from the EM optimization if $\sum_k p(\omega_{i'} | \boldsymbol{x}_k) = 0$. In the reduced problem, we would be guaranteed that $\sum_k p(\omega_{i} | \boldsymbol{x}_k) > 0$ $\forall i$, which (as we showed above) ensures that the EM parameters always lie in the interior of the parameter space.

\subsection{Proving Differentiability of $Q(q(\boldsymbol{\omega}) ; q^{(s)}(\boldsymbol{\omega}) )$ over $(\boldsymbol{\Omega} \setminus \partial\boldsymbol{\Omega}) \bigtimes \boldsymbol{\Omega}$}
\label{sec:proving_smoothness_of_Q}

Given our assumption that $p(\omega_i) \geq \epsilon $ $\forall i$, we will show that $Q(q(\boldsymbol{\omega}) ; q^{(s)}(\boldsymbol{\omega}) )$ is not only continuous but also differentiable over $(\boldsymbol{\Omega} \setminus \partial\boldsymbol{\Omega}) \bigtimes \boldsymbol{\Omega}$. Because $Q(q(\boldsymbol{\omega}) ; q^{(s)}(\boldsymbol{\omega}) )$ is a composition of differentiable functions, to prove differentiability over $(\boldsymbol{\Omega} \setminus \partial\boldsymbol{\Omega}) \bigtimes \boldsymbol{\Omega}$ it suffices to show that $Q(q(\boldsymbol{\omega}) ; q^{(s)}(\boldsymbol{\omega}) )$ is defined over $(\boldsymbol{\Omega} \setminus \partial\boldsymbol{\Omega}) \bigtimes \boldsymbol{\Omega}$.

Building on the expression for $Q(q(\boldsymbol{\omega}), q^{(s)}(\boldsymbol{\omega}))$ from \textbf{Eqn. \ref{eqn:definition_of_Q}}, we have:
\begin{align}
    Q(q(\boldsymbol{\omega}), q^{(s)}(\boldsymbol{\omega})) &= \sum_{k=1}^{N}\sum_{i=1}^{m}q^{(s)}(\omega_i|\boldsymbol{x}_k)\left(\log q(\omega_i) + \log p(\boldsymbol{x}_k|\omega_i) \right) \nonumber\\
    &= \sum_{k=1}^{N}\sum_{i=1}^{m}q^{(s)}(\omega_i|\boldsymbol{x}_k) \left( \log q(\omega_i) + \log \left[\frac{p(\omega_i|\boldsymbol{x}_k)p(\boldsymbol{x}_k)}{p(\omega_i)}\right] \right) \label{eqn:applybayesruletoq}\\
    &= \sum_{k=1}^{N}\sum_{i=1}^{m}\frac{q^{(s)}(\omega_i)p(\omega_i | \boldsymbol{x}_k)\left( \log q(\omega_i) + \log \left[\frac{p(\omega_i|\boldsymbol{x}_k)p(\boldsymbol{x}_k)}{p(\omega_i)}\right] \right)}{p(\omega_i) \sum_{j=1}^m \left( \frac{q^{(s)}(\omega_j)}{p(\omega_j)} p(\omega_j | \boldsymbol{x}_k)\right) } \label{eqn:subpsomegagivenx}\\
    &= \sum_{k=1}^{N}\sum_{i=1}^{m}\frac{q^{(s)}(\omega_i)p(\omega_i | \boldsymbol{x}_k)\left(\log q(\omega_i) + \log p(\omega_i|\boldsymbol{x}_k) + \log p(\boldsymbol{x}_k) - \log p(\omega_i) \right)}{p(\omega_i) \sum_{j=1}^m \left( \frac{q^{(s)}(\omega_j)}{p(\omega_j)} p(\omega_j | \boldsymbol{x}_k)\right) } \nonumber\\
    &= \sum_{k=1}^{N}\sum_{i=1}^{m}\frac{q^{(s)}(\omega_i) p(\omega_i | \boldsymbol{x}_k)\left(\log q(\omega_i) + \log p(\boldsymbol{x}_k) - \log p(\omega_i) \right) + q^{(s)}(\omega_i)\log \left[ p(\omega_i|\boldsymbol{x}_k)^{p(\omega_i|\boldsymbol{x}_k)} \right]}{p(\omega_i) \sum_{j=1}^m \left( \frac{q^{(s)}(\omega_j)}{p(\omega_j)} p(\omega_j | \boldsymbol{x}_k)\right) } \label{eqn:rewrittenQ}
\end{align}

Where (\ref{eqn:applybayesruletoq}) comes from applying Bayes' rule to $p(\boldsymbol{x}_k | \omega_i)$, and (\ref{eqn:subpsomegagivenx}) follows from substituting the definition of $q^{(s)}(\omega_i | \boldsymbol{x}_k)$ from \textbf{Eqn. \ref{eqn:qs_omega_given_x}}.

We make two observations: first, because $x^x > 0$ for all $x \ge 0$, the term $\log \left[p(\omega | \boldsymbol{x}_k)^{p(\omega | \boldsymbol{x}_k)}\right]$ is always defined. Second, because $\log p(\boldsymbol{x_k})$ does not depend on $q(\omega_i)$, it disappears when optimizing $Q(q(\boldsymbol{\omega}), q^{(s)}(\boldsymbol{\omega}))$ w.r.t $q(\omega_i)$. If we assume, as before, that $p(\omega_i) \ge  \epsilon$ $\forall i$, we see that $Q(q(\boldsymbol{\omega}), q^{(s)}(\boldsymbol{\omega}))$ is defined if and only if $q(\omega_i) > 0$ $\forall i$ and $\sum_{j=1}^m \frac{q^{(s)}(\omega_j)}{ p(\omega_j) } p(\omega_j | \boldsymbol{x}_k)$ $\forall k$. The first condition holds true when $q(\boldsymbol{\omega}) \in (\boldsymbol{\Omega} \setminus \partial \boldsymbol{\Omega})$, and the second condition holds true for all $q^{(s)}(\boldsymbol{\omega}) \in \boldsymbol{\Omega}$. Thus, we conclude that the domain of $Q(q(\boldsymbol{\omega}), q^{(s)}(\boldsymbol{\omega}))$ is $(\boldsymbol{\Omega} \setminus \partial\boldsymbol{\Omega}) \bigtimes \boldsymbol{\Omega}$. As stated earlier, because $Q(q(\boldsymbol{\omega}), q^{(s)}(\boldsymbol{\omega}))$ is composed of differentiable functions, it follows that $Q(q(\boldsymbol{\omega}), q^{(s)}(\boldsymbol{\omega}))$ is both continuous and differentiable over $(\boldsymbol{\Omega} \setminus \partial\boldsymbol{\Omega}) \bigtimes \boldsymbol{\Omega}$.

\subsection{Invoking Theorem 1 of Wu (1983) to Prove EM Converges to a Stationary Point of the Likelihood}
\label{sec:invoke_theorem_1}

We now show that EM converges to a stationary point of the log-likelihood. We will closely follow the derivation of Theorem 2 in Wu (1983), but modifying it so as not to assume that $Q(q(\boldsymbol{\omega}), q^{(s)}(\boldsymbol{\omega}))$ is defined for all $q(\boldsymbol{\omega}) \in \boldsymbol{\Omega}$. Our proof will leverage Theorem 1 of Wu (1983), which is reproduced below for convenience:

\textbf{Theorem 1 of Wu (1983)}: Let $\boldsymbol{\Omega}$ denote the domain of the likelihood function $L(\Phi)$, let $\mathscr{S}$ denote the set of stationary points in the interior of $\boldsymbol{\Omega}$, and let $\{ \Phi_p \}$ denote a sequence of Generalized EM (GEM) parameter updates generated by $\Phi_{p+1} \in M(\Phi_p)$. If (i) $M$ is a closed point-to-set map over the complement of $\mathscr{S}$ and (ii) $L(\phi_{p+1}) > L(\phi)$ for all $\phi_p \notin \mathscr{S}$. Then all the limit points of $\{ \Phi_p \}$ are stationary points of $L$, and $L(\Phi_p)$ converges monotonically to $L^* = L(\Phi^*)$ for some $\Phi^* \in \mathscr{S}$. \QED

To prove convergence to a stationary point, it suffices to show that both (i) and (ii) of Theorem 1 apply to the case of domain adaptation to label shift. We verify (i) and (ii) in turn below.

\subsubsection{(i) of Theorem 1: $M$ is a closed point-to-set map over the complement of $\mathscr{S}$}
We first state what it means for $M$ to be a closed point-to-set map. A map $M$ from points of $X$ to subsets of $X$ is called a point-to-set map on $X$. In our case, $M$ is a point-to-point map (a special case of a point-to-set map) given by the EM parameter update. A point-to-point map is said to be \emph{closed} at $x$ if $x_k \rightarrow x$, $x \in X$ and $y_k \rightarrow y$, $y_k = A(x_k)$ implies $y = A(x)$. For a point-to-point map, continuity implies closedness. Because the complement of $\mathscr{S}$ corresponds to the domain $\boldsymbol{\Omega} \setminus \mathscr{S}$, if we show that $M$ is continuous over the entirety of $\boldsymbol{\Omega}$, we would satisfy our conditions.

Referring to \textbf{Eqn. \ref{eqn:em_update_rule}}, we see that $M$ is defined as:
\begin{align*}
    M(q^{(s)}(\boldsymbol{\omega}))_i &= \frac{1}{N} \sum_{k=1}^N q^{(s)} (\omega_i | \boldsymbol{x}_k)\\
    &= \frac{1}{N} \sum_{k=1}^N \frac{\frac{q^{(s)}(\omega_i)}{p(\omega_i)}p(\omega_i|\boldsymbol{x}_k)}{\sum_{j=1}^{m}\frac{q^{(s)}(\omega_j)}{p(\omega_j)}p(\omega_j|\boldsymbol{x}_k)}
\end{align*}

Because $M$ is composed of differentiable functions, to show continuity of $M$ over $\boldsymbol{\Omega}$ it suffices to show that $M(q^{(s)}(\boldsymbol{\omega}))$ is defined for all $q^{(s)}(\boldsymbol{\omega}) \in \boldsymbol{\Omega}$. We observe that, given our assumption of $p(\omega_i) \ge \epsilon$ $\forall i$, the numerator $\frac{q^{(s)}(\omega_i)}{p(\omega_i)}$ is always defined. In order for the denominator $ \sum_{j=1}^{m}\frac{q^{(s)}(\omega_j)}{p(\omega_j)}p(\omega_j|\boldsymbol{x}_k)$ to be defined $\forall k$, we additionally require that $\sum_{j} q^{(s)}(\omega_j) p(\omega_j | \boldsymbol{x}_k) > 0$ $\forall k$. The latter condition is satisfied if $q^{(s)}(\boldsymbol{\omega}) \in \boldsymbol{\Omega}$. Thus, we conclude that $M$ is continuous over $\boldsymbol{\Omega}$, fulfilling (i) of Theorem 1 of Wu (1983). \QED

\subsubsection{(ii) of Theorem 1: $L(\phi_{p+1}) > L(\phi)$ for all $\phi_p \notin \mathscr{S}$}

We first show that $\frac{\partial H(q(\boldsymbol{\omega}); q^{(s)}(\boldsymbol{\omega}) )}{ \partial q(\boldsymbol{\omega})} = 0$ when $q(\boldsymbol{\omega}) = q^{(s)}(\boldsymbol{\omega})$. For notational convenience, we will use $D^{10} H(a;b)$ and $D^{10} Q(a;b)$ to denote $\frac{\partial H(\Phi_1; \Phi_2 )}{ \partial \Phi_1}$ and $\frac{\partial Q(\Phi_1; \Phi_2 )}{ \partial \Phi_1}$ at $\Phi_1 = a$ and $\Phi_2 = b$, and will use $DL(a)$ to denote $\frac{\partial L(\Phi)}{\partial \Phi}$ at $\Phi=a$.

In \textbf{Sec. \ref{sec:proving_smoothness_of_Q}}, we established that $Q(q(\boldsymbol{\omega}), q^{(s)}(\boldsymbol{\omega}) )$ is differentiable for all $q(\boldsymbol{\omega}) \in (\boldsymbol{\Omega} \setminus \partial \boldsymbol{\Omega})$. Because $H(q(\boldsymbol{\omega}), q^{(s)}(\boldsymbol{\omega})) := Q(q(\boldsymbol{\omega}), q^{(s)}(\boldsymbol{\omega})) - L(q(\boldsymbol{\omega}))$, it follows that $H(q(\boldsymbol{\omega}), q^{(s)}(\boldsymbol{\omega}))$ is also differentiable for all $q(\boldsymbol{\omega}) \in (\boldsymbol{\Omega} \setminus \partial \boldsymbol{\Omega})$. In \textbf{Sec. \ref{sec:qs_interior_of_omega}}, we established that $q^{(s)}(\boldsymbol{\omega}) \in (\boldsymbol{\Omega} \setminus \partial \boldsymbol{\Omega})$ $\forall s$; thus, it follows that $H(q(\boldsymbol{\omega}), q^{(s)}(\boldsymbol{\omega}))$ is differentiable in the neighborhood of $q^{(s)}(\boldsymbol{\omega})$. Because $q(\boldsymbol{\omega})=q^{(s)}(\boldsymbol{\omega})$ maximizes $H(q(\boldsymbol{\omega}), q^{(s)}(\boldsymbol{\omega}))$ (\textbf{Eqn. \ref{eqn:qsw_maximises_H}}), we conclude that $D^{10} H(q^{(s)}(\boldsymbol{\omega}); q^{(s)}(\boldsymbol{\omega})) = 0$.

We now show that $D^{10} H(q^{(s)}(\boldsymbol{\omega}); q^{(s)}(\boldsymbol{\omega})) = 0$ and $q^{(s)}(\boldsymbol{\omega}) \notin \mathscr{S}$ implies $Q(q^{(s+1)}(\boldsymbol{\omega}); q^{(s)}(\boldsymbol{\omega})) > Q(q^{(s)}(\boldsymbol{\omega}); q^{(s)}(\boldsymbol{\omega}))$. From the definition of $H(q^{(s)}(\boldsymbol{\omega}), q^{(s)}(\boldsymbol{\omega}))$, we have $L(q^{(s)}(\boldsymbol{\omega})) = Q(q^{(s)}(\boldsymbol{\omega}), q^{(s)}(\boldsymbol{\omega})) - H(q^{(s)}(\boldsymbol{\omega}), q^{(s)}(\boldsymbol{\omega}))$. If $D^{10}H(q^{(s)}(\boldsymbol{\omega}); q^{(s)}(\boldsymbol{\omega})) = 0$ and $q^{(s)}(\boldsymbol{\omega}) \notin \mathscr{S}$, we have $DL(q^{(s)}(\boldsymbol{\omega})) = D^{10}Q(q^{(s)}(\boldsymbol{\omega}); q^{(s)}(\boldsymbol{\omega})) \ne 0$. Because $q^{(s)}(\boldsymbol{\omega}) \in (\boldsymbol{\Omega} \setminus \partial \boldsymbol{\Omega})$ (\textbf{Sec. \ref{sec:qs_interior_of_omega}}) and $Q(q(\boldsymbol{\omega}), q^{(s)}(\boldsymbol{\omega}))$ is both defined and differentiable $\forall q(\boldsymbol{\omega}) \in (\boldsymbol{\Omega} \setminus \partial \boldsymbol{\Omega})$ (\textbf{Sec. \ref{sec:proving_smoothness_of_Q}}), we conclude that if $D^{10}Q(q^{(s)}(\boldsymbol{\omega}); q^{(s)}(\boldsymbol{\omega})) \ne 0$, then $q(\boldsymbol{\omega}) = q^{(s)}(\boldsymbol{\omega})$ does not maximize $Q(q(\boldsymbol{\omega}), q^{(s)}(\boldsymbol{\omega}))$. Because $q^{(s+1)}(\boldsymbol{\omega}) = \argmax_{q(\boldsymbol{\omega})} Q(q(\boldsymbol{\omega}), q^{(s)}(\boldsymbol{\omega}))$ by definition, we get $Q(q^{(s+1)}(\boldsymbol{\omega}); q^{(s)}(\boldsymbol{\omega})) > Q(q^{(s)}(\boldsymbol{\omega}); q^{(s)}(\boldsymbol{\omega}))$ when $q^{(s)}(\boldsymbol{\omega}) \notin \mathscr{S}$.

Finally, we show that if $Q(q^{(s+1)}(\boldsymbol{\omega}); q^{(s)}(\boldsymbol{\omega})) > Q(q^{(s)}(\boldsymbol{\omega}); q^{(s)}(\boldsymbol{\omega}))$, then $L(q^{(s+1)}(\boldsymbol{\omega})) > L(q^{(s)}(\boldsymbol{\omega}))$. Analogous to the derivation of \textbf{Eqn. \ref{eqn:em_monotonic_improvement}}, we have:

\begin{align*}
    L(q^{(s+1)}(\boldsymbol{\omega})) &= Q(q^{(s+1)}(\boldsymbol{\omega}); q^{(s)}(\boldsymbol{\omega})) - H(q^{(s+1)}(\boldsymbol{\omega}); q^{(s)}(\boldsymbol{\omega})) \text{ (From the definition of }H\text{)}\nonumber\\
    &\text{ If } Q(q^{(s+1)}(\boldsymbol{\omega}); q^{(s)}(\boldsymbol{\omega})) > Q(q^{(s)}(\boldsymbol{\omega}); q^{(s)}(\boldsymbol{\omega})) \text{, we get:}\\
    L(q^{(s+1)}(\boldsymbol{\omega})) &> Q(q^{(s)}(\boldsymbol{\omega}); q^{(s)}(\boldsymbol{\omega})) - H(q^{(s+1)}(\boldsymbol{\omega}); q^{(s)}(\boldsymbol{\omega})) \nonumber\\
    &\ge Q(q^{(s)}(\boldsymbol{\omega}); q^{(s)}(\boldsymbol{\omega})) - H(q^{(s)}(\boldsymbol{\omega}); q^{(s)}(\boldsymbol{\omega})) \text{ (From \textbf{Eqn. \ref{eqn:qsw_maximises_H}})} \nonumber\\
    &= L(q^{(s)}(\boldsymbol{\omega}))
\end{align*}

Thus, we have shown that $L(q^{(s+1)}(\boldsymbol{\omega})) > L(q^{(s)}(\boldsymbol{\omega}))$ $\forall q^{(s)}(\boldsymbol{\omega}) \notin \mathscr{S}$, fulfilling (ii) of Theorem 1 of Wu (1983).

\QED

\end{appendix}

\end{document}